\documentclass[11pt]{article}

\usepackage[utf8]{inputenc}
\usepackage[T1]{fontenc}
\usepackage{lmodern}
\usepackage[margin=1in]{geometry}
\usepackage{graphicx}
\usepackage{amsmath,amssymb}
\usepackage{booktabs}
\usepackage{array}
\usepackage{multirow}
\usepackage{caption}
\usepackage{subcaption}
\usepackage{xcolor}
\usepackage{enumitem}
\usepackage{authblk}
\usepackage[hidelinks,colorlinks=true,linkcolor=blue,citecolor=blue,urlcolor=blue]{hyperref}
\usepackage{natbib}
\usepackage{microtype}
\usepackage{textcomp}

\title{\textbf{A Computer Vision Pipeline for Individual-Level Behavior Analysis: Benchmarking on the Edinburgh Pig Dataset}}

\author[1]{H. Yang}
\author[1]{E. Liu}
\author[1]{J. Sun}
\author[1]{S. Sharma}
\author[1]{M. van Leerdam}
\author[2]{S. Franceschini}
\author[1]{P. Niu}
\author[1]{M. Hostens}

\affil[1]{Department of Animal Science, College of Agriculture and Life Sciences, Cornell University, Ithaca, NY 14853, USA}
\affil[2]{University of Li\`ege, Gembloux Agro-Bio Tech (UL\`iege-GxABT), 5030 Gembloux, Belgium}

\date{}

\begin{document}

\maketitle

\begin{abstract}
Animal behavior analysis plays a crucial role in understanding animal welfare, health status, and productivity in agricultural settings. However, traditional manual observation methods are time-consuming, subjective, and limited in scalability. We present a modular pipeline that leverages open-sourced state-of-the-art computer vision techniques to automate animal behavior analysis in a group housing environment. Our approach combines state-of-the-art models for zero-shot object detection, motion-aware segmentation and tracking, and advanced feature extraction using vision transformers for robust behavior recognition. The pipeline addresses challenges including animal occlusions and group housing scenarios, as demonstrated in indoor pig monitoring. We validated our system on the Edinburgh Pig Behavior Video Dataset for multiple behavioral tasks. Our temporal model achieved 94.2\% overall accuracy, representing a 21.2 percentage point improvement over existing methods. The pipeline demonstrated robust tracking capabilities with a 93.3\% identity preservation (IDF1) score and an 89.3\% average precision (AP) for object detection. The modular design suggests potential for adaptation to other contexts, though further validation across species would be required. The open-source implementation provides a scalable solution for behavior monitoring, contributing to precision pig farming and welfare assessment through automated, objective, and continuous analysis.

\vspace{0.5em}
\noindent\textbf{Highlights:}
\begin{itemize}[leftmargin=*,topsep=2pt,itemsep=0pt]
    \item Modular pipeline achieves 94.2\% accuracy in recognition of nine different behaviors in pigs.
    \item Real-time tracking maintains 93.3\% identity preservation in group housing.
    \item Open-source pipeline processes video to behavioral classification end-to-end.
\end{itemize}

\vspace{0.5em}
\end{abstract}

\section{Introduction}

Welfare is an increasing concern in the livestock industry. \citet{cornish2016} highlighted growing public awareness, and \citet{clark2019} documented increased regulations regarding farm animal conditions. In modern agricultural practice, measuring animal behavior has become increasingly important for ensuring optimal welfare, productivity, and health management \citep{berckmans2014, halachmi2019}. Animal behavior serves as a vital indicator of various physiological and psychological states \citep{bugueiro2021}. \citet{matthews2016} demonstrated that automated detection of behavioral changes in pigs could identify health and welfare compromises before clinical manifestation, and further found that changes in activity patterns and feeding behavior could predict health issues days before traditional clinical diagnosis. Similarly, \citet{antanaitis2023} showed how monitoring rumination, eating, and locomotion behaviors using sensors could assess cattle responses to heat stress, while \citet{dzermeikaite2023} discussed how continuous behavioral monitoring through artificial intelligence and machine learning enables early disease detection in cattle farming. Behavioral monitoring often provides early warning signs of disease, stress, or environmental discomfort before clinical symptoms become apparent.

However, as \citet{neethirajan2021} noted, traditional monitoring methods, which rely heavily on human observation, are labor-intensive, prone to subjective interpretation, and limited in both temporal coverage and scalability. Similarly, \citet{matthews2017} emphasized that continuous observation of livestock by farm staff is impractical in commercial settings to the degree required for detecting behavioral changes relevant for early intervention. \citet{hampton2019} further indicated that sample sizes required for reasonable levels of precision in animal welfare assessments often exceed 300 animals, which is typically unfeasible for traditional observation-based methods due to the time, labor, and logistical constraints involved. These limitations have created a pressing need for automated, objective, and continuous monitoring solutions.

While sensor-based monitoring has shown benefits for health detection and improved outcomes \citep{firk2002, rial2024}, these systems face limitations including low farmer confidence and inability to assess complex social behaviors crucial for welfare assessment \citep{eckelkamp2020, stygar2021}. Computer vision and artificial intelligence have emerged as promising alternatives, offering advantages such as eliminating physical attachments, reducing labor demands, and providing continuous behavioral monitoring at lower costs \citep{oliveira2021, tian2020, mcdonagh2021}. However, deploying computer vision in real farm environments presents unique challenges beyond typical applications \citep{menezes2024}.

Traditional livestock farming presents several complex challenges for computer vision systems. First, farm infrastructure creates severe occlusions through fences, feeding equipment, and overlapping animals with similar appearances \citep{li2021}. Second, dramatic lighting variations from natural and artificial sources can decrease model performance by 20--30\% without adaptation \citep{wurtz2019, fuentes2023}. Third, camera mounting constraints often result in suboptimal viewing angles limited to partial top-down or side views \citep{psota2020}. Fourth, group housing causes frequent occlusions and identity switches, as demonstrated by \citet{guo2023} who documented multiple bounding box switching events across 759 videos. Finally, individual-level identification must handle challenges like odd poses and appearance changes \citep{vidal2021}.

\citet{rohan2024} highlighted how recent advances in deep learning, particularly in object detection, tracking, and feature extraction, have opened new possibilities for addressing these challenges. Behavior detection in animals using video analysis involves a multi-step computational process. A video is treated as a sequence of consecutive images, where the first step is to locate the target animal in the initial frame. Once detected, the same animal must be tracked across subsequent frames to ensure consistency throughout the analysis. The animal is then cropped from each frame, removing redundant information such as the background and other animals. This series of cropped images forms the input for feature extraction, where visual information, such as shape, movement, or pixel intensity, is mathematically represented. These features serve as inputs to a classification model. Finally, this classification model is trained using an annotated dataset, also known as ground truth, enabling it to recognize key behaviors, such as distinguishing a lying cow from a standing one.

For object detection, open-vocabulary models like OWLv2 \citep{minderer2023} enable zero-shot detection of animals without requiring species-specific training, offering advantages over traditional architectures in agricultural settings. Video tracking has been transformed by the progression from SAM \citep{kirillov2023} to SAM\,2 \citep{ravi2024} and most recently SAMURAI \citep{yang2024}, which incorporates motion-aware memory and Kalman filtering for robust tracking in dynamic farm environments. Feature extraction has evolved from manual descriptors to self-supervised learning approaches, with DINOv2 \citep{oquab2023} learning high-quality visual features without annotations and CLIP \citep{radford2021} enabling zero-shot classification through vision-language alignment---particularly valuable when labeled agricultural data is scarce. These advances enable behavior classification models to achieve high accuracy, as demonstrated by \citet{domun2019} who achieved 95\% accuracy for pig behavior recognition, though challenges remain in handling group dynamics and individual tracking in complex farm settings.

Despite these advances, \citet{rohan2024} observed that more recent deep learning approaches, while showing promise, typically focus on specific tasks or controlled environments, lacking the flexibility and robustness required for comprehensive behavior analysis in real-world agricultural settings. \citet{bonneau2020} found that even advanced hybrid systems combining deep learning and time-lapse cameras for outdoor animal monitoring achieve sensitivity rates ranging from only 70.7\% to 94.8\%, with performance varying dramatically based on environmental factors. \citet{liu2024} concluded that conventional animal tracking methods consistently fail to meet the precision and real-time speed requirements necessary for practical application due to persistent challenges including occlusion, complex backgrounds, and identification switches. Most critically, while individual components have shown promise in isolation, there remains a lack of integrated pipelines that combine these various techniques into a cohesive, end-to-end solution for specific agricultural applications. This motivates the development of modular approaches that can generate usable feature representations on a frame level from videos, though such pipelines require validation for each specific use case.

To address these limitations, we propose a modular pipeline that integrates multiple open-sourced state-of-the-art computer vision techniques into a cohesive pipeline. Our pipeline consists of six main components: (1)~optimized video decoding for efficient frame processing to overcome the temporal coverage limitations of traditional methods that \citet{hampton2019} identified as requiring large sample sizes; (2)~zero-shot object detection using OWLv2 (or YOLOv12 as needed) for initial animal localization that addresses the poor accuracy of conventional systems in farm environments highlighted by \citet{fernandes2020}; (3)~motion-aware segmentation and tracking with the SAMURAI model from \citet{yang2024} for continuous individual monitoring to overcome the specific challenges of occlusion and animal overlapping documented by \citet{guo2023}; (4)~automated object cropping for isolation of individual animals, which mitigates the identification difficulties in group housing described by \citet{vidal2021}; (5)~feature extraction using DINOv2 or CLIP for robust representation learning that can handle the variable lighting conditions and performance decreases of 20--30\% noted by \citet{fuentes2023}; and (6)~flexible classification architectures for behavior recognition. Aside from the pipeline architecture itself, we tried camera footage from top-mounted cameras as a potential solution to mitigate the farm's intricate layouts mentioned by \citet{li2021}, even though they only include partial information \citep{psota2020}.

In short, we demonstrate how open-sourced state-of-the-art algorithms can be integrated into a modular pipeline for individual-level behavior analysis, validated specifically on pig behavior recognition in group housing environments.

\section{Materials and Methods}

\subsection{Datasets Description}

We deployed our pipeline on two open-sourced datasets: (1)~the CBVD-5 dataset from \citet{li2024} and (2)~the Edinburgh Pig Behavior Video Dataset from \citet{bergamini2021}. As we only validated the feature extraction ability of the model on the CBVD-5 dataset, the details of that experiment are presented in the Supplementary Material.

\subsubsection{Edinburgh Pig Behavior Video Dataset}

The Edinburgh Pig Behavior Video Dataset from \citet{bergamini2021} represents a comprehensive video dataset specifically designed for automated pig behavior recognition and welfare monitoring research. The dataset captures focused video recordings from a nearly overhead perspective of a single pen housing with eight growing pigs. Videos were recorded in RGB color format at six frames per second with a resolution of $1280\times720$ pixels, providing sufficient temporal and spatial resolution for behavior analysis while maintaining manageable data volumes. The pen environment featured standard commercial pig farming infrastructure designed to reflect typical industry conditions. This included a three-space feeder and two nipple water drinkers, with flooring consisting of partially slatted surfaces supplemented with straw and shredded paper bedding. Such a realistic setup ensures the dataset's relevance for developing automated livestock monitoring systems that can be deployed in an actual commercial farming setting. The ground truth annotations were meticulously prepared to provide comprehensive information for each visible pig in every labeled frame. These annotations included three key elements: axis-aligned bounding boxes that precisely delineate each animal's spatial location; persistent tracking identifiers that maintain individual pig identity across consecutive frames; and behavior labels categorized into 17 distinct predefined classes. The dataset comprises 7{,}200 annotated frames, with eight pigs tracked in each frame, providing approximately 57{,}600 individual pig annotations. This substantial volume of detailed annotations offered a robust foundation for training and evaluating computer vision algorithms across multiple tasks, including pig detection and localization, individual tracking in group housing environments, and behavior recognition at the individual level. The combination of detailed annotations and consistent labeling protocols made this dataset particularly suitable for thoroughly evaluating the comprehensive capabilities of our proposed pipeline.

For evaluation, nine video sequences (600 frames each) were utilized for tracking performance assessment and 800 frames were randomly sampled from the complete dataset for object detection benchmarking. This comprehensive annotation scheme enabled us to evaluate all components of our pipeline: detection, tracking, and behavior classification.

\subsection{Computing Environment}

Our experiments were conducted on a computing cluster featuring an NVIDIA V100 GPU (16\,GB memory), 6 core CPUs, 112\,GB RAM, and a local NVMe SSD array provided in Databricks. We implemented the pipeline using PyTorch 2.0, with additional libraries including OpenCV for video processing and scikit-learn for evaluation metrics. The selection of a GPU cluster is mainly due to the hardware requirements of the SAMURAI tracking model. For other sections in the pipeline, a CPU cluster would be efficient for all data manipulation.

\subsection{System Architecture Overview}

\begin{figure}[!htbp]
    \centering
    \includegraphics[width=\textwidth]{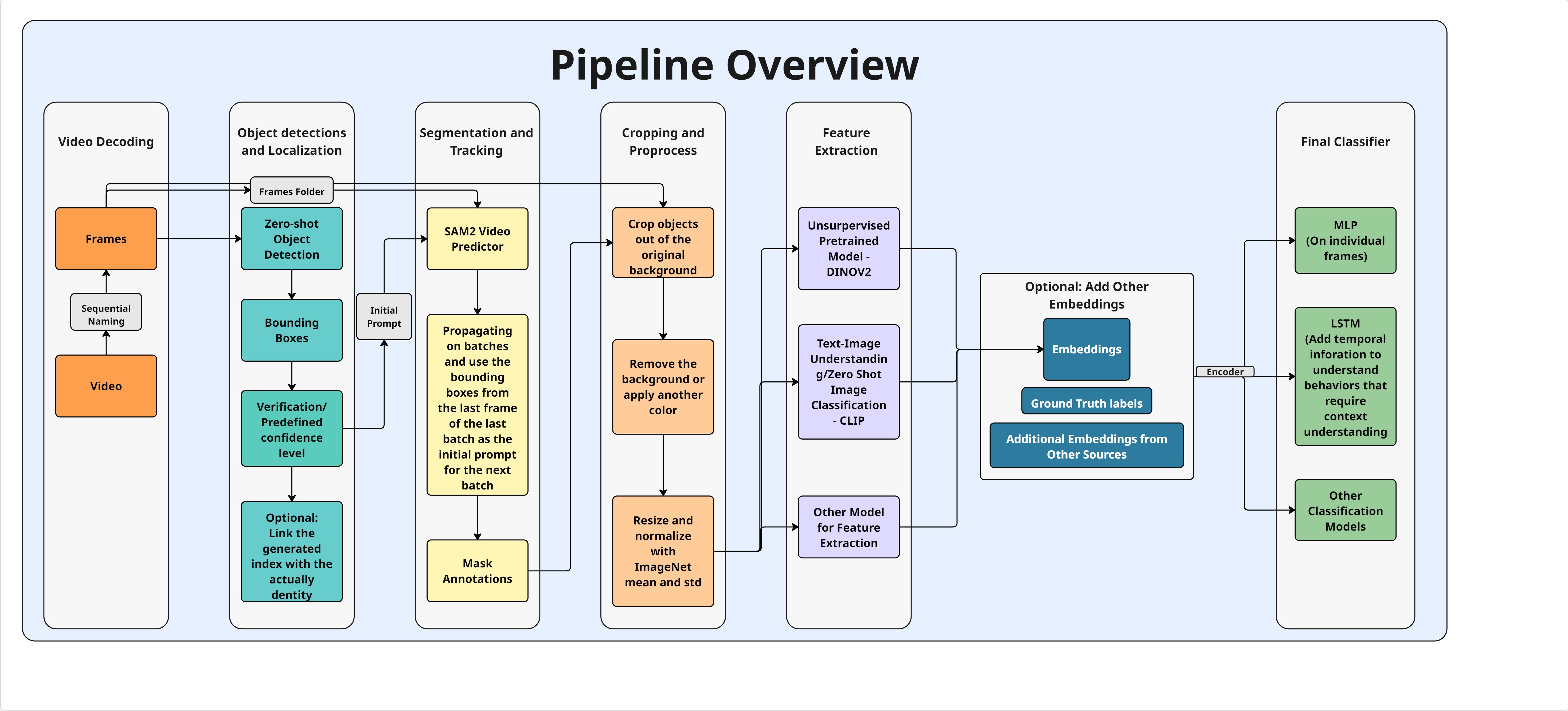}
    \caption{System architecture overview showing the complete workflow from video decoding to behavior classification: raw video is decoded into sequential frames, and objects are localized to seed segmentation and temporal tracking with the SAM\,2 video predictor. Each detected instance is then cropped (background removed or recolored), resized, and normalized before feature extraction using unsupervised pretrained models (e.g., DINOv2) or zero-shot classifiers (e.g., CLIP). Finally, per-frame embeddings are fed to an MLP for instantaneous behavior labels, while sequences of embeddings are passed through an LSTM (or other temporal model) to capture behavioral dynamics.}
    \label{fig:pipeline}
\end{figure}

Our framework is designed as a modular pipeline that processes video input through six distinct stages: video decoding, object detection, segmentation and tracking, object cropping, feature extraction, and behavior classification, as shown in Figure~\ref{fig:pipeline}. All those stages will be explained in the following sections.

The pipeline begins with raw video input from farm cameras, demonstrated here with overhead-mounted cameras in pig housing, and then goes through our framework:
\begin{enumerate}[leftmargin=*,topsep=2pt,itemsep=2pt]
    \item The videos are first decoded into sequential frames with optimized sampling to balance temporal resolution with computational efficiency.
    \item The decoded frames then undergo object detection using OWLv2 or YOLOv12 to identify and localize individual animals in the first frame.
    \item Once initial detections are established, the SAMURAI model performs continuous tracking and segmentation throughout the video sequence, producing precise masks and bounding boxes for each animal across all frames.
    \item The tracked objects are then cropped from their original frames, creating individual image sequences for each animal.
    \item These cropped sequences undergo feature extraction using either DINOv2 or CLIP, generating rich embeddings that capture both visual appearance and behavioral patterns.
    \item Finally, these embeddings serve as input to various classification architectures, from simple MLPs to sophisticated temporal models, depending on the specific behavior analysis requirements.
\end{enumerate}

This modular design ensures flexibility and allows for easy replacement or upgrading of individual components as new technologies emerge.

\subsection{Video Decoding and Frame Organization}

The video decoding process is critical for managing computational resources while maintaining sufficient temporal resolution for behavior analysis. Our approach implements several optimizations to handle the large data volumes typical in continuous farm monitoring.

Videos are processed using OpenCV's efficient video capture capabilities, with configurable stride parameters to control frame sampling density. The stride value determines how many frames are skipped between saved frames, allowing us to adjust the temporal resolution based on the specific behaviors being analyzed. For instance, fast movements like jumping require higher temporal resolution (lower stride values), while behaviors like walking can be adequately captured with lower temporal resolution (higher stride values).

To prevent Graphics Processing Unit (GPU) memory overflow during the tracking phase, we organize decoded frames into subfolders with a maximum of 3{,}000 frames each. This limit was determined empirically through extensive testing on various GPU configurations. When tracking four objects simultaneously, we found that processing more than 5{,}000 frames would consistently cause memory issues on standard GPUs like the NVIDIA V100 with 16\,GB memory.

A global sequential numbering scheme (e.g., \texttt{0000001.jpg}, \texttt{0000002.jpg}, \ldots) was employed for naming image frames, ensuring temporal continuity across subfolders. This is crucial for preserving the temporal relationships needed for behavior analysis, especially when behaviors span multiple subfolders.

\subsection{Zero-Shot Object Detection}

\subsubsection{OWLv2}

OWLv2 \citep{minderer2023} is an open-vocabulary object detector that supports zero-shot detection via text-based queries, making it well-suited to agricultural environments with variable object categories. Building on OWL-ViT (Open-World Localization Vision Transformer) \citep{minderer2022}, OWLv2 introduces a self-training approach (OWL-ST) that uses pseudo-box annotations generated from over a billion web images and their associated text, enabling the model to learn at web scale without manual annotations. OWLv2 also introduces several architectural optimizations that further enhance efficiency (removing low-variance image patches) and improve instance selection using an object head, reducing computation by approximately 50\% compared to the original OWL-ViT. After object detection, a two-stage filtering process was applied: (1)~automatic filtering based on target object classes (e.g., ``pig'' in our implementation) and confidence thresholds defined according to the specific application, where objects not meeting these criteria were excluded; and (2)~manual refinement through bounding box overlays (Figure~\ref{fig:cattle_detection}), where the model's output was examined visually. The evaluation employed an Intersection over Union (IoU) threshold of 0.5 to assess detection quality.

We also experimented with the YOLOv12 model \citep{tian2025}. However, in our experiments with the Edinburgh Pig Dataset, OWLv2 proved more effective than YOLOv12 for detecting pigs in overhead views, demonstrating the importance of selecting appropriate models for specific applications.

\subsection{Motion-Aware Segmentation and Tracking with SAMURAI}

Segmentation and tracking is an essential step that generates sequential images of the same animal chronologically. SAM\,2, the successor of SAM, enables promptable video segmentation. SAMURAI builds upon the foundation of SAM\,2, introducing critical enhancements for video tracking in agricultural settings. Its core innovation is a motion-aware approach that integrates Kalman filter-based motion modeling with selective memory mechanisms.

Compared to the vanilla SAM\,2 video detector, SAMURAI demonstrates substantial performance improvements across multiple challenging benchmarks. On the LaSOT dataset \citep{fan2019}, which encompasses 70 object categories including livestock, amphibians, reptiles, arthropods, and other mammals across more than 3.87 million frames, SAMURAI achieves up to 5.69\% AUC gain and 6.53\% $P_{\text{norm}}$ gain, making it particularly well-suited for long-term single-object tracking applications. The model further exhibits a 7.1\% AUC improvement on LaSOText \citep{fan2021}, showcasing its robustness in handling diverse tracking scenarios. Additionally, SAMURAI achieves a 3.5\% AO gain on the GOT-10k benchmark \citep{huang2019}, which features 563 object classes and 87 distinct motion patterns, demonstrating its effectiveness in generic, class-agnostic tracking tasks. These consistent improvements across varied benchmarks establish SAMURAI as the current state-of-the-art video tracking model, offering superior performance for applications requiring robust object tracking in complex, real-world scenarios.

To handle long video sequences, a batch processing strategy with memory management was implemented. After each subfolder of frames was processed, the final bounding boxes were extracted to serve as initial prompts for the next batch. GPU memory was cleared between batches using garbage collection and PyTorch's memory management functions to prevent the accumulation of memory fragmentation.

To evaluate tracking performance, the MOTA Challenge metric was used to comprehensively assess tracking accuracy, identity preservation, and trajectory continuity.

\subsection{Automated Object Cropping}

The cropping process isolates individual animals from their background, creating standardized inputs for feature extraction. This step is crucial for ensuring consistent feature quality regardless of the animal's position or size in the original frame.

Our implementation uses parallel processing with \texttt{ProcessPoolExecutor}, a high-level Python API for running callables in a pool of separate processes rather than threads. Each cropping task processes a single frame--annotation pair through the following steps: (1)~load the original frame using OpenCV; (2)~extract the bounding box coordinates $[x, y, w, h]$ from the annotation; (3)~create a binary mask using the contour information; (4)~apply the mask to isolate the object from the background; (5)~fill external regions with a specified background color (typically black or white); and (6)~resize the cropped region to a standard dimension (e.g., $224\times224$ pixels).

The output filename convention preserves traceability: \texttt{[global\_frame\_index]\_[object\_name].jpg}. This naming scheme maintains the temporal sequence while identifying individual animals, essential for subsequent temporal analysis.

Parallel processing significantly improves throughput, with the number of concurrent workers configurable based on available CPU cores. However, we implement safeguards to prevent system overload by leaving 2--3 cores idle to avoid the disk I/O bottleneck, and we monitor system load to adjust the worker count dynamically when necessary.

\subsection{Feature Extraction}

Feature extraction was performed on cropped images to generate high-dimensional embeddings using two complementary approaches: DINOv2 for visual features and Contrastive Language-Image Pre-training (CLIP) for multimodal representations. Both approaches produce embeddings well-suited to downstream behavior analysis. The performance of CLIP versus DINOv2 was compared using the CBVD-5 dataset; details are provided in the Supplementary Material.

\subsubsection{DINOv2 Architecture and Implementation}

DINOv2 \citep{oquab2023} is a self-supervised vision model that learns powerful visual features without requiring labeled data. Starting with 1.2B uncurated images, DINOv2 uses deduplication and self-supervised retrieval to create a curated dataset of 142M diverse images (LVD-142M). The model is trained efficiently using techniques like FlashAttention and sequence packing, with resolution adaptation in the final training phase.

For smaller models, DINOv2 employs knowledge distillation from the largest ViT-g model. The result is a family of models (ViT-S/B/L/g) that achieve state-of-the-art performance on various vision tasks without fine-tuning, making them excellent general-purpose visual encoders for both image-level and pixel-level applications.

Our implementation uses DINOv2-large, which offers an optimal balance between accuracy and computational efficiency. The model processes each cropped image through the following pipeline: (1)~image preprocessing with standard normalization; (2)~forward pass through the Vision Transformer; (3)~extraction of the \texttt{[CLS]} token representation; (4)~optional mean pooling over spatial dimensions; and (5)~saving the resulting embedding as a \texttt{.pt} file.

The resulting embedding has shape $(1, 1024)$, capturing rich, learned representations of the cropped images that can be fed into a behavior-classification model.

\subsection{Behavior Classification}

In order to translate the generated embeddings into behavior classifications of the nine different pig behaviors, two different deep learning models were applied. The model performance was compared using the evaluation metrics: precision, recall, F1-score, and support.

\subsubsection{Basic MLP Classifier}

For simple behavior classification tasks, we implement a straightforward Multi-Layer Perceptron (MLP) architecture to evaluate the pipeline's performance even with basic classification models. The architecture consists of a three-layer feedforward neural network that processes the extracted feature embeddings. The input layer accepts the embedding vectors from our feature extraction stage, which are 1024-dimensional when using DINOv2-large. The first hidden layer reduces this dimensionality to 512 neurons using a fully connected transformation, followed by ReLU activation to introduce non-linearity and a dropout rate of 0.5 for regularization. The second hidden layer further compresses the representation to 256 neurons, maintaining the same ReLU activation and 0.5 dropout rate to prevent overfitting during training. Finally, the output layer maps these features to the number of target behavior classes through a linear transformation followed by softmax activation, producing normalized probability distributions over the possible behaviors.

\subsubsection{Temporal Models}

For behaviors requiring temporal context, we implement a Bidirectional LSTM (BiLSTM) architecture that captures both forward and backward temporal dependencies. The model consists of a single-layer bidirectional LSTM with 128 hidden units per direction, resulting in a 256-dimensional output when concatenating both forward and backward states. This BiLSTM layer processes sequences of 1024-dimensional DINOv2 embeddings extracted from consecutive frames, enabling the model to learn temporal patterns that are crucial for behavior recognition.

The architecture employs a classification head consisting of two fully connected layers. The first linear layer reduces the 256-dimensional BiLSTM output to 128 neurons, followed by a ReLU activation function and dropout regularization with a rate of 0.3 to prevent overfitting. The second linear layer maps these features to the final number of behavior classes. We extract the hidden state from the last timestep of the sequence as the comprehensive temporal representation, which encapsulates the accumulated behavioral patterns throughout the observed time window.

\subsubsection{Model Training and Evaluation}

The MLP classifier was trained using the backpropagation algorithm \citep{rumelhart1986} with the Adam optimizer \citep{kingma2014}, employing a learning rate of $1\times10^{-3}$ and weight decay of $1\times10^{-5}$ for regularization. To address class imbalance in the dataset, we computed class weights using inverse frequency weighting, where each class weight was calculated as the inverse of its frequency normalized by the sum of all inverse frequencies. These weights were incorporated into the cross-entropy loss function to ensure balanced learning across all behavior categories.

The dataset was split using stratified sampling to maintain class distributions, with 70\% for training, 15\% for validation, and 15\% for testing. Training proceeded for a maximum of 50 epochs with early stopping based on validation loss, using a patience of 10 epochs to prevent overfitting. The model achieving the lowest validation loss was saved and used for final evaluation. All experiments were conducted on NVIDIA GPUs with a batch size of 64 and 4 parallel data loading workers to optimize computational efficiency.

\subsection{Modular Architecture Design}

Our pipeline is designed with a modular architecture to address the inherent complexity and diversity of animal behavior analysis in agricultural settings. The modular approach follows principles established in software engineering \citep{bass2021} and computer vision systems \citep{orhei2020} that promote component isolation, independent development, and flexible reconfiguration.

The six core modules (decoding, detection, tracking, cropping, feature extraction, and classification) are designed with well-defined interfaces that facilitate: (1)~independent optimization, where each module can be separately refined or replaced without disrupting the entire pipeline; (2)~flexible deployment, where different subsets of modules can be deployed based on specific research or application requirements; and (3)~incremental improvement, where new techniques can be incorporated into specific modules as they become available.

\section{Results}

\subsection{Validation on the Edinburgh Pig Behavior Video Dataset}

\subsubsection{Dataset Preparation}

\begin{figure}[!htbp]
    \centering
    \includegraphics[width=\textwidth]{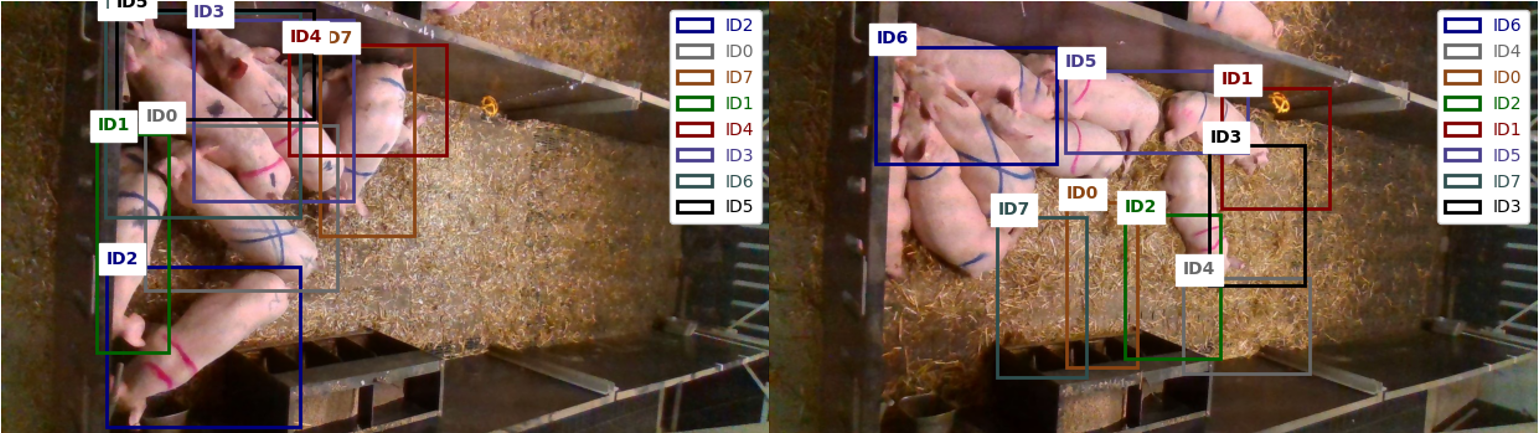}
    \caption{Examples of sequences excluded from benchmarking: (left)~dataset showing misalignment between ground truth bounding boxes and actual pig locations, and (right)~challenging tracking scenario with invisible objects on the first frame due to severe overlapping and mounting behaviors in the corner.}
    \label{fig:excluded}
\end{figure}

We decoded the 12 annotated video sequences using the stride specified on the dataset's official website, following the procedure described in Section~2.4. This produced 600 frames per sequence, accumulating 7{,}200 labeled frames with 8 labeled pigs each, for a total of 57{,}600 individual pig annotations. The annotations were verified by overlaying the bounding boxes on the first frame, and two issues were discovered. First, in one sequence the initial annotations did not align with the objects and that sequence was therefore excluded from later benchmarking (Figure~\ref{fig:excluded}, left). Second, in two other sequences the starting frame contained pigs mounting one another in a corner, with some pigs not visible from the camera, and these sequences were also excluded (Figure~\ref{fig:excluded}, right). The remaining 9 sequences were used for thorough benchmarking, yielding 43{,}200 labeled individuals.

\subsubsection{Object Detection Benchmarking}

Visual comparison between YOLO and OWLv2 revealed that YOLO could not generate predictions for all the objects present in the frame without fine-tuning on the pig dataset, contrary to our earlier findings on the cow dataset. We therefore selected OWLv2, which performed reliably out of the box. The detailed comparison is provided in the Supplementary Material.

In our experiments, all pigs were detected at a confidence threshold of 0.5, which we adopted for validation on the decoded frames.

At the standard IoU threshold of 0.50, the model achieved an Average Precision (AP) of 89.28\%, demonstrating robust detection capabilities. As shown in Table~\ref{tab:detection}, the system maintained a favorable balance between precision (80.19\%) and recall (88.05\%), resulting in an F1 score of 83.94\%. The true positive rate of 88.05\% indicates that the model successfully detected the vast majority of pigs, while maintaining a relatively low false positive rate of 19.81\%. The mean IoU of correctly matched detections was 0.747, suggesting accurate localization beyond the minimum threshold requirement.

The model's counting accuracy yielded a Mean Absolute Error (MAE) of 1.53 pigs per frame, indicating reasonable performance in estimating the number of animals present in each image. Counting accuracy is a critical metric for livestock monitoring applications where accurate population assessment is essential.

\begin{table}[!htbp]
    \centering
    \caption{Object detection performance metrics for OWLv2 on the Edinburgh Pig Behavior dataset at IoU threshold 0.50.}
    \label{tab:detection}
    \begin{tabular}{lc}
        \toprule
        \textbf{Metric} & \textbf{Value} \\
        \midrule
        Average Precision (AP) & 89.28\% \\
        Precision              & 80.19\% \\
        Recall                 & 88.05\% \\
        F1 Score               & 83.94\% \\
        True Positive Rate     & 88.05\% \\
        False Positive Rate    & 19.81\% \\
        Missed Detection Rate  & 11.95\% \\
        Average IoU            & 0.747 \\
        \bottomrule
    \end{tabular}
\end{table}

\subsubsection{Object Segmentation and Tracking Benchmarking}

The proposed tracking system achieved an average Multiple Object Tracking Accuracy (MOTA) of 86.68\%. The IDF1 score, which measures the system's ability to correctly identify and maintain object identities throughout sequences, reached 93.33\%. The system exhibited remarkable consistency in identity management, with an average of only 0.44 identity switches across all sequences. Track fragmentations averaged 83.89, with the number of tracklets matching the actual number of pigs in each sequence, while maintaining a perfect average tracklet length of 600 frames. This demonstrates the system's ability to maintain continuous tracks throughout entire video sequences without interruption.

\begin{figure}[!htbp]
    \centering
    \includegraphics[width=\textwidth]{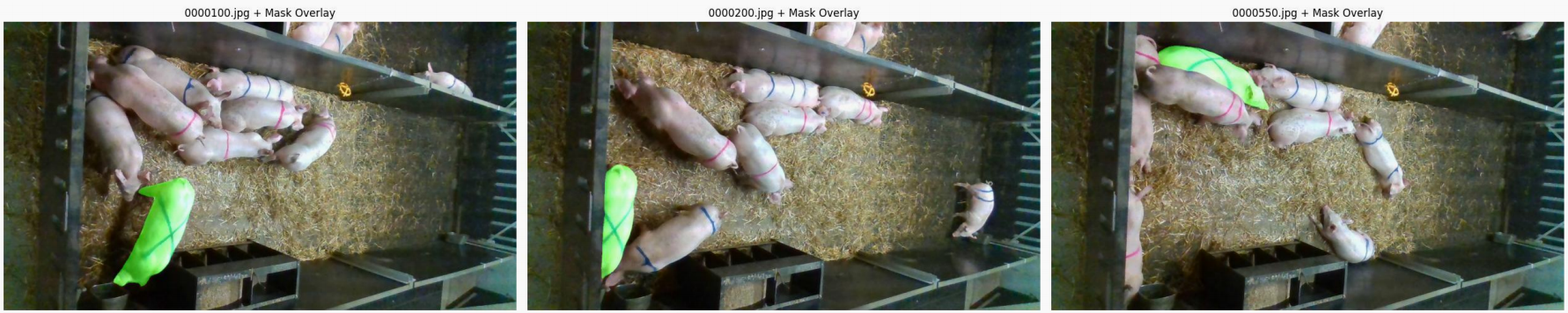}
    \caption{Visualization example of SAMURAI tracking performance demonstrating robust identity preservation during severe occlusion events. The green mask represents the object being tracked. Across different timestamps (time evolving from left to right), the mask identity of the pig stays the same even when the tracked object overlaps with other entities from the view of the camera.}
    \label{fig:samurai_tracking}
\end{figure}

Individual sequence analysis revealed consistently high performance across diverse scenarios. The system achieved IDF1 scores ranging from 86.40\% to 99.90\% as shown in Table~\ref{tab:tracking}, with five sequences exceeding 90\% identity preservation. The best performance was observed in sequence 2019\_12\_10\_000060, which achieved 99.90\% IDF1 and 99.80\% MOTA with only 5 missed detections and no identity switches, as shown in Figure~\ref{fig:samurai_tracking}.

\begin{table}[!htbp]
    \centering
    \caption{Multi-object tracking performance metrics across nine validation sequences from the Edinburgh Pig Behavior dataset.}
    \label{tab:tracking}
    \small
    \setlength{\tabcolsep}{4pt}
    \begin{tabular}{lcccccccc}
        \toprule
        \textbf{Validation Sequence} & \textbf{IDF1} & \textbf{Recall} & \textbf{Precision} & \textbf{ML} & \textbf{Switches} & \textbf{MOTA} & \textbf{Avg.\ Length} & \textbf{Tracklets} \\
        \midrule
        2019\_11\_05\_000002 & 91.00\% & 91.00\% & 91.00\% & 0 & 0    & 82.00\% & 600 & 8 \\
        2019\_11\_11\_000028 & 87.40\% & 87.40\% & 87.40\% & 0 & 2    & 74.80\% & 600 & 8 \\
        2019\_11\_11\_000036 & 91.60\% & 91.60\% & 91.60\% & 0 & 0    & 83.20\% & 600 & 8 \\
        2019\_11\_22\_000010 & 86.40\% & 86.40\% & 86.40\% & 0 & 0    & 72.80\% & 600 & 8 \\
        2019\_11\_28\_000113 & 98.30\% & 98.30\% & 98.30\% & 0 & 0    & 96.60\% & 600 & 8 \\
        2019\_12\_02\_000005 & 94.50\% & 94.50\% & 94.50\% & 0 & 0    & 89.10\% & 600 & 8 \\
        2019\_12\_02\_000208 & 98.10\% & 98.10\% & 98.10\% & 0 & 0    & 96.20\% & 600 & 8 \\
        2019\_12\_10\_000060 & 99.90\% & 99.90\% & 99.90\% & 0 & 0    & 99.80\% & 600 & 8 \\
        2019\_12\_10\_000078 & 92.80\% & 92.80\% & 92.80\% & 0 & 2    & 85.60\% & 600 & 8 \\
        \midrule
        \textbf{Average}     & \textbf{93.33\%} & \textbf{93.33\%} & \textbf{93.33\%} & \textbf{0} & \textbf{0.44} & \textbf{86.68\%} & \textbf{600} & \textbf{8} \\
        \bottomrule
    \end{tabular}
    \vspace{2pt}
    \begin{flushleft}\footnotesize ML: Mostly Lost. Switches: Number of identity switches. Avg.\ Length: Average tracklet length (frames).\end{flushleft}
\end{table}

The tracking system's precision and recall both averaged 93.33\%, indicating balanced performance in detecting true positives while minimizing false detections. The consistency of these metrics across sequences suggests reliable performance suitable for automated behavioral monitoring applications. The complete elimination of tracklet fragmentation in the majority of sequences, with all animals tracked as ``mostly tracked'' or ``partially tracked'' and none classified as ``mostly lost,'' further validates the system's effectiveness for continuous monitoring applications in precision pig farming.

\subsubsection{Feature Extraction and MLP Classification Model Benchmarking}

We first cropped pig regions from the background using the bounding box annotations, following the procedure described in Section~2.7, and then extracted high-dimensional visual embeddings from these cropped frames for use in behavior classification.

The feature extraction process used parallel processing across 12 workers to efficiently handle the computational demands of the DINOv2-large model. Because the dataset annotations propagate forward in time, we cropped each object using its current descriptor until a new descriptor superseded it, yielding a total of 43{,}200 unique behavioral instances across 16 distinct behavior categories. Each cropped pig image, standardized to $224\times224$ pixels, was processed through the pre-trained transformer to generate 1024-dimensional feature vectors. These embeddings captured rich visual representations suitable for distinguishing between behavioral patterns.

The distribution of behaviors in our dataset revealed significant class imbalance, with `investigating' (10{,}281 instances) and `walk' (2{,}766 instances) being the most frequent, while rare behaviors such as `chase' (1 instance) and `jumpontopof' (6 instances) had insufficient samples for reliable classification. To ensure robust model training and evaluation, we selected nine behaviors with adequate representation: standing (3{,}168), lying (3{,}187), eating (5{,}475), drinking (638), sitting (327), sleeping (15{,}256), running (90), playing with toy (126), and nose-to-nose interactions (431).

To visualize the learned feature space, we employed t-SNE dimensionality reduction on a subset of embeddings representing nine key behaviors: drinking, eating, lying, nose-to-nose, playing with toy, running, sitting, sleeping, and standing. The resulting visualization revealed distinct clustering patterns, with certain behaviors forming well-separated groups while others showed expected overlap due to visual similarities. For instance, stationary behaviors such as eating and drinking formed cohesive clusters, while dynamic behaviors like running and playing with toy exhibited more dispersed distributions in the feature space.

\begin{figure}[!htbp]
    \centering
    \includegraphics[width=0.75\textwidth]{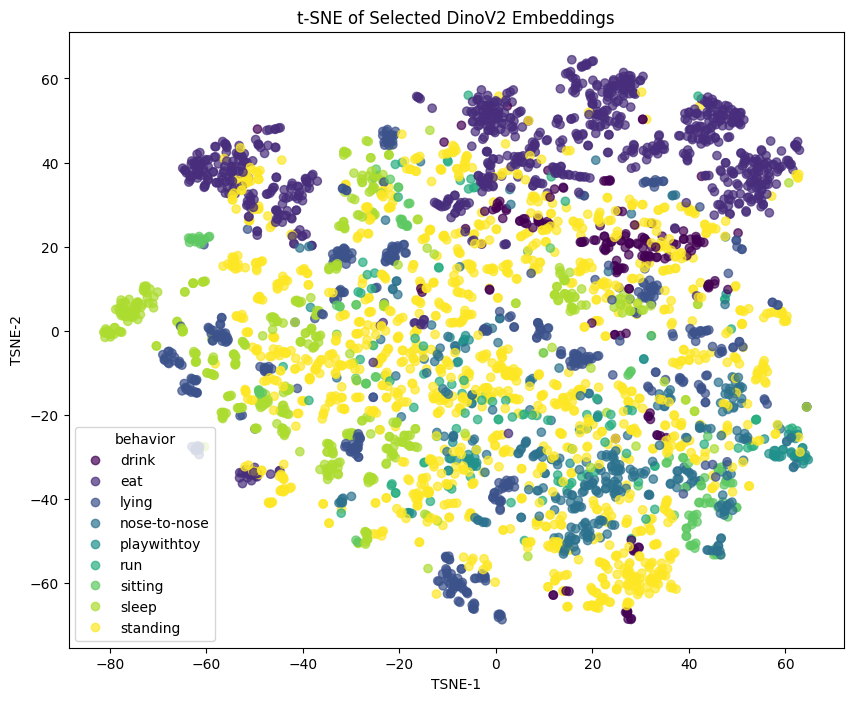}
    \caption{t-SNE visualization of DINOv2 embeddings revealing natural clustering of pig behaviors with distinct separation between stationary (eating, sleeping) and dynamic (running, playing) activities.}
    \label{fig:tsne_dinov2}
\end{figure}

\paragraph{(1) MLP Classification Results}

We first evaluated a Multi-Layer Perceptron (MLP) classifier on the extracted DINOv2 features. After filtering to nine well-represented behaviors (standing, lying, eating, drinking, sitting, sleeping, running, playing with toy, and nose-to-nose interactions), we obtained 28{,}698 examples with a 70/15/15 train/validation/test split.

The MLP classifier achieved a test accuracy of 92.9\%, as shown in Table~\ref{tab:mlp}. The model demonstrated high performance across most behavior categories, with eating achieving the highest F1-score (0.980), followed by sleeping (0.964) and lying (0.883). The system showed particularly strong recall for eating (99.6\%), lying (96.2\%), and nose-to-nose interactions (93.8\%), indicating reliable detection when these behaviors occurred.

The confusion matrix in Table~\ref{tab:mlp} also revealed that while standing behavior achieved high precision (89.2\%), it showed moderate recall (76.2\%), with misclassifications distributed across lying, eating, and nose-to-nose interactions. Dynamic behaviors like running presented challenges, achieving 64.3\% recall but lower precision (47.4\%), with confusion primarily occurring with standing and nose-to-nose interactions.

\paragraph{(2) LSTM Classification Results}

To better capture temporal dependencies in behavioral sequences, we implemented an LSTM-based classifier that processes sequences of DINOv2 embeddings. Using a sliding window approach with majority-based filtering, we generated 14{,}255 temporal windows from the original dataset.

The LSTM classifier achieved a test accuracy of 94.2\%, representing a 1.3 percentage point improvement over the MLP baseline as shown in Table~\ref{tab:lstm}. This temporal model demonstrated superior performance in several key areas. Notably, the LSTM achieved more balanced precision-recall trade-offs across behaviors, with weighted average metrics of 94.9\% precision, 94.3\% recall, and 94.4\% F1-score.

The LSTM showed improvements in challenging behavior categories. Standing behavior maintained high precision (90.7\%) while achieving 78.4\% recall, an improvement over the MLP's 76.2\%. The model demonstrated exceptional performance on eating (97.2\% F1-score) and sleeping (97.8\% F1-score). Social behaviors such as nose-to-nose interactions achieved 90.3\% recall with the LSTM, though precision remained moderate at 51.7\%; this suggests that temporal context helped identify these interactions, but distinguishing them from other close-proximity behaviors remained challenging.

Sitting behavior showed lower recall with the LSTM (76.0\%) compared to the MLP (87.8\%), and also lower precision, resulting in a lower overall F1-score (66.7\% vs.\ 75.4\%). This suggests the temporal model was more conservative in identifying sitting postures.

\begin{table}[!htbp]
    \centering
    \caption{Per-class classification performance of MLP model with DINOv2 features.}
    \label{tab:mlp}
    \begin{tabular}{lcccc}
        \toprule
        \textbf{Behavior} & \textbf{Precision} & \textbf{Recall} & \textbf{F1-Score} & \textbf{Support} \\
        \midrule
        Standing          & 0.892 & 0.762 & 0.821 & 475 \\
        Lying             & 0.816 & 0.962 & 0.883 & 478 \\
        Eating            & 0.965 & 0.996 & 0.980 & 821 \\
        Drinking          & 0.860 & 0.896 & 0.878 & 96 \\
        Sitting           & 0.662 & 0.878 & 0.754 & 49 \\
        Sleeping          & 0.992 & 0.937 & 0.964 & 2289 \\
        Running           & 0.473 & 0.643 & 0.546 & 14 \\
        Playing with toy  & 0.900 & 0.947 & 0.923 & 19 \\
        Nose-to-nose      & 0.492 & 0.938 & 0.645 & 64 \\
        \midrule
        \textbf{Weighted Average} & \textbf{0.940} & \textbf{0.929} & \textbf{0.932} & \textbf{4305} \\
        \bottomrule
    \end{tabular}
\end{table}

\begin{table}[!htbp]
    \centering
    \caption{Per-class classification performance of LSTM model with DINOv2 features.}
    \label{tab:lstm}
    \begin{tabular}{lcccc}
        \toprule
        \textbf{Behavior} & \textbf{Precision} & \textbf{Recall} & \textbf{F1-Score} & \textbf{Support} \\
        \midrule
        Standing          & 0.907 & 0.784 & 0.841 & 236 \\
        Lying             & 0.901 & 0.958 & 0.929 & 238 \\
        Eating            & 0.985 & 0.958 & 0.972 & 409 \\
        Drinking          & 0.793 & 0.958 & 0.868 & 48 \\
        Sitting           & 0.594 & 0.760 & 0.667 & 25 \\
        Sleeping          & 0.983 & 0.973 & 0.978 & 1136 \\
        Running           & 0.400 & 0.571 & 0.471 & 7 \\
        Playing with toy  & 0.818 & 1.000 & 0.900 & 9 \\
        Nose-to-nose      & 0.517 & 0.903 & 0.700 & 31 \\
        \midrule
        \textbf{Weighted Average} & \textbf{0.949} & \textbf{0.943} & \textbf{0.944} & \textbf{2139} \\
        \bottomrule
    \end{tabular}
\end{table}

\section{Discussion}

\subsection{Comparison with Former Research}

\subsubsection{Benchmarking Results Comparison}

Our object detection results using OWLv2 achieved an AP of 89.28\%, which is 5.93 percentage points lower than the 95.21\% AP reported by \citet{bergamini2021} using a fine-tuned YOLOv3 model. However, direct comparison is challenging as the details about which sequences were chosen as the validation set were not disclosed in the original study. The difference in performance may be attributed to the zero-shot nature of our approach versus their fine-tuned model, as well as potential differences in validation set composition.

For tracking performance, our pipeline demonstrated substantial improvements over the baseline model reported by \citet{bergamini2021}. The MOTA increased from 84.88\% to 86.68\%, representing a modest but meaningful improvement. More significantly, the IDF1 score improved from 71.78\% to 93.33\%, a 21.55 percentage point increase that indicates superior identity preservation capabilities. The reduction in identity switches from 16 to 0.44 (97.2\% reduction) and decrease in track fragmentations from 115 to 83.89 demonstrate the effectiveness of our approach. To ensure a fair comparison given the 600-frame constraint of our annotated sequences, we specifically benchmarked our results against validation sequences A and D from \citet{bergamini2021}, which exhibit comparable average tracklet lengths.

In behavior classification, our framework demonstrated substantial improvements over previous approaches. The MLP classifier achieved 92.9\% accuracy on nine behavior categories, while the LSTM classifier further improved performance to 94.2\%. Both models significantly outperformed the 73\% average accuracy reported by \citet{bergamini2021} on five behaviors (standing, lying, moving, eating, and drinking).

This represents a 19.9 percentage point improvement for MLP and 21.2 percentage point improvement for LSTM over the baseline, despite evaluating on a more diverse and challenging set of nine behaviors. The superior performance can be attributed to several factors: (1)~the use of modern vision transformer features (DINOv2) that capture richer visual representations compared to traditional CNN features; (2)~the effectiveness of our preprocessing pipeline that ensures high-quality individual animal crops; and (3)~for the LSTM model, the incorporation of temporal context that captures behavioral dynamics.

The LSTM's advantage over MLP was particularly evident in behaviors with temporal characteristics. While both models achieved similar performance on static behaviors like lying (MLP: 88.3\% F1, LSTM: 92.9\% F1), the LSTM showed marked improvements for dynamic and transitional behaviors. The temporal modeling also improved the precision-recall balance, as evidenced by the weighted average F1-score improvement from 93.2\% (MLP) to 94.4\% (LSTM).

Notably, even our simpler MLP architecture substantially exceeded the original baseline, suggesting that modern pre-trained vision transformers like DINOv2 can effectively encode behavioral information without requiring complex temporal modeling for many applications. However, LSTM's consistent improvements across most behavior categories validate the importance of temporal context for comprehensive behavior analysis.

These results demonstrate that our modular pipeline, combining state-of-the-art vision models with appropriate architectural choices, can achieve high accuracy in automated behavior classification while handling increased behavioral complexity. These improvements were achieved specifically on pig behavior analysis, and similar gains would need to be validated for other species or agricultural contexts.

\subsection{Practical Benefits of the Modular Design}

The modular design provides three concrete benefits. First, processing-intensive operations can be optimized independently: the decoding and tracking modules include specialized memory management to handle long video sequences, while feature extraction employs parallel processing to maximize throughput. Resource-constrained environments can also implement only a subset of the pipeline---for example, deploying only the feature extraction and classification modules on pre-recorded video when real-time processing is not required. Second, errors in one module do not necessarily cascade through the system; if tracking temporarily fails due to occlusion, downstream stages can recover in subsequent frames. Third, individual modules can be replaced as conditions or technologies change, as demonstrated when we switched from YOLOv12 to OWLv2 for pig detection without redesigning the rest of the pipeline. This same property suggests potential for adaptation to other species and environments, though, as our YOLOv12 experience shows, each new context requires empirical validation rather than guaranteed transfer.

\subsection{Limitations}

\subsubsection{High-Quality Initial Frames and Strict Data Flow Conventions}

The pipeline requires that animals in the first frame are not severely overlapping, mounting, or missing, as these conditions prevent successful initial detection and compromise the entire downstream process. Additionally, the modular design requires strict adherence to standardized data flow conventions, which users must follow precisely for proper functionality.

\subsubsection{Camera Positioning Trade-offs}

Camera positioning represents a critical design decision with inherent trade-offs that significantly impact system performance. Side-view camera configurations provide superior visibility of limb movements and postural details, enabling more nuanced behavioral classification; however, they suffer from frequent occlusions when monitoring multiple animals, potentially compromising tracking continuity. Conversely, top-view camera installations substantially reduce inter-animal occlusions and simplify instance segmentation but significantly limit access to limb information that \citet{mathis2018} demonstrated to be essential for accurate behavior detection and classification. Multi-camera systems that integrate both perspectives offer comprehensive coverage and redundancy, theoretically overcoming the limitations of single-view approaches; however, this solution introduces substantially higher system complexity in terms of hardware requirements, calibration procedures, and computational demands for data fusion, alongside proportionally increased implementation costs that may limit practical deployment in commercial agricultural settings.

\subsubsection{Computational Trade-offs}

Higher image resolutions improve detection accuracy but incur quadratic increases in computational demands. Real-time processing, while valuable for immediate interventions, requires architectural compromises that may reduce accuracy compared to offline analysis. These trade-offs necessitate careful optimization based on specific deployment requirements and available resources.

\subsubsection{Species-Specific Validation Requirements}

Our experience with YOLOv12 failing on pig detection while OWLv2 succeeded highlights that model selection remains species-specific. Despite using state-of-the-art ``zero-shot'' models, each new application context requires careful validation and potentially different model choices, limiting immediate generalizability.

\subsection{Future Work}

Several extensions to the method itself merit investigation. Multi-modal integration could augment visual features with audio (for vocalizations that lack distinct visual signatures), environmental sensors (for context-aware behavioral analysis), or genomic data for individual-specific modeling \citep{alvarenga2021}, integrated through hierarchical fusion networks with weight-aware modules that dynamically balance modality importance \citep{wang2024, bokade2021}. However, multi-modal integration must be approached judiciously, as additional modalities can introduce redundant information that diminishes effectiveness \citep{liu2024, arablouei2023}, and each new data source requires empirical evaluation. Targeted fine-tuning of pretrained components such as SAM-2-Video---which was pre-trained on roughly 35.5 million mask instances, far beyond what most labs can produce---is also possible, but only when paired with careful curation of a compact, diverse annotation set and freezing of most pretrained weights to safeguard the knowledge acquired during large-scale pre-training. Beyond pigs, the zero-shot capabilities of OWLv2 and the general-purpose design of SAMURAI suggest applicability to other species, though our YOLOv12 experience shows that systematic evaluation across species, housing conditions, and camera configurations is essential.

On the deployment side, current computational requirements limit use in resource-constrained farm environments. Knowledge distillation, post-training quantization, and structured pruning could yield more compact models suitable for edge devices, while streaming pipelines and distributed processing architectures could enable real-time analysis on standard farm surveillance systems. Practical adoption will also depend on integration with existing farm management infrastructure: classification outputs can be formatted for common farm platforms, and the embeddings produced by our pipeline are highly storage-efficient---a typical 2\,GB video compresses to roughly 0.05\,GB of numerical representations while preserving behavioral information, easing data sharing and meta-analyses across studies \citep{wurtz2019}. To support these directions, our pipeline will be released as an open-source project with standardized configuration files, consistent naming conventions, and example configurations for common scenarios. We encourage the research community to contribute improvements and report performance on new species and environments.

\section{Conclusions and Perspectives}

We have presented a modular pipeline for automated behavior analysis validated on pig monitoring in group housing environments. By integrating state-of-the-art deep learning techniques including OWLv2 for detection, SAMURAI for tracking, and DINOv2 for feature extraction, our pipeline achieved 94.2\% accuracy on nine-class pig behavior recognition using temporal models.

Our experiments on the Edinburgh Pig Behavior Video Dataset demonstrated substantial improvements over existing methods, including a 21.2 percentage point increase in classification accuracy and a 21.55 percentage point improvement in identity preservation (IDF1) during tracking. The modular architecture enabled us to adapt to specific challenges---such as switching from YOLOv12 to OWLv2 when the former failed on pig detection---highlighting both the flexibility of the design and the need for empirical validation in each application context.

While our results on pig behavior analysis are promising, several limitations must be acknowledged. The pipeline requires high-quality initial frames for successful detection, and even state-of-the-art ``zero-shot'' models required species-specific selection. These findings underscore that validation on additional species and environments would be necessary before broader claims about generalizability can be made.

This work demonstrates how existing computer vision models can be effectively integrated for livestock behavior analysis when properly validated and configured. The open-source implementation provides researchers with a tested framework for pig behavior monitoring and a potential starting point for adaptation to other contexts. As precision livestock farming continues to evolve, such automated monitoring tools---when properly validated for each specific application---can contribute to improved animal welfare assessment and management decisions.

\appendix
\renewcommand{\thesection}{S\arabic{section}}
\setcounter{section}{0}
\renewcommand{\thefigure}{S\arabic{figure}}
\setcounter{figure}{0}
\renewcommand{\thetable}{S\arabic{table}}
\setcounter{table}{0}

\section*{Supplementary Material}
\addcontentsline{toc}{section}{Supplementary Material}

\section{Considerations and Trials for Deciding Model}

In order to identify the best model, we conducted some evaluations of several open-source models. Two distinct video datasets were used: a proprietary recording of our own dairy cows and the publicly available Edinburgh Pig Behavior Video \citep{bergamini2021}.

\subsection{Object Detection and Localization}

\begin{figure}[!htbp]
    \centering
    \includegraphics[width=\textwidth]{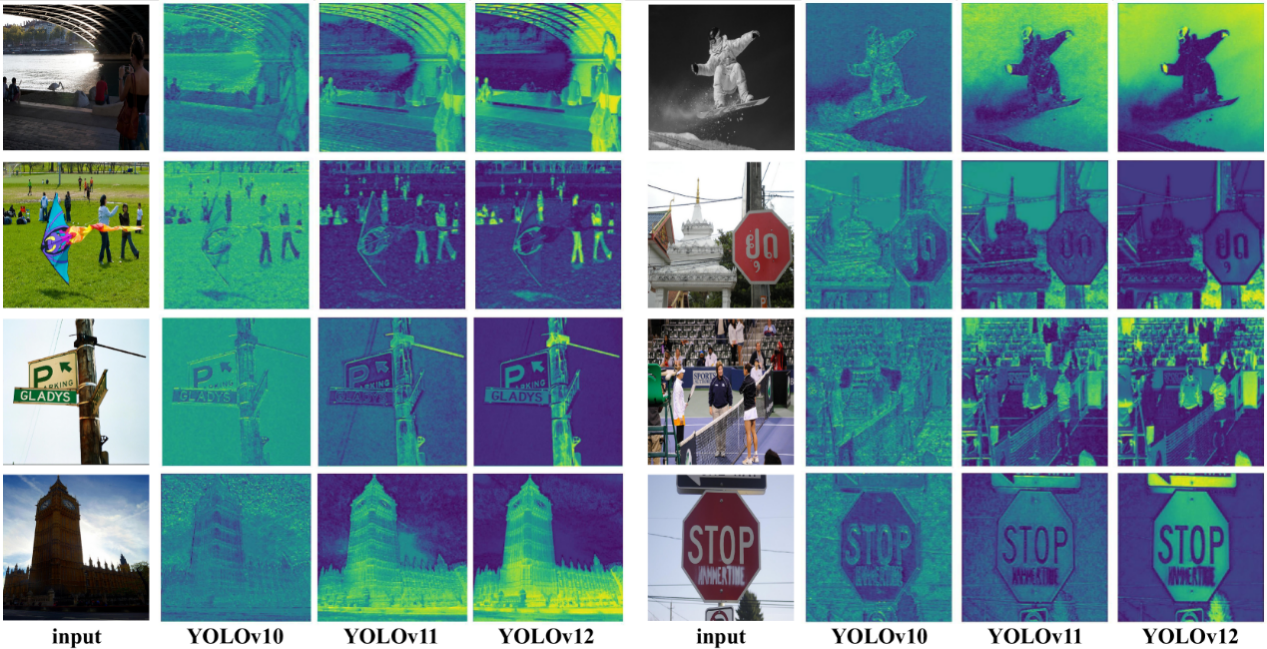}
    \caption{Comparative visualization of attention heat maps across YOLO versions, demonstrating YOLOv12's superior object localization and background suppression \citep{tian2025}. YOLOv12 exhibits a better ability to extract object contours from the same input images.}
    \label{fig:yolo_attention}
\end{figure}

Aside from the OWLv2 model, we also experimented with the YOLOv12 model. YOLOv12 represents a significant advancement in the YOLO family by integrating an innovative area attention (A2) mechanism that overcomes the computational limitations of traditional attention approaches, which typically scale quadratically with input size and become impractical for high-resolution agricultural imagery \citep{tian2025}. By dividing feature maps into distinct areas and computing attention locally, A2 reduces complexity to linear time, enabling efficient processing of large images while preserving global context. This attention-centric, real-time object detection framework outperforms previous YOLO versions and other detectors across all model scales (N, S, M, L, X). YOLOv12-L achieves a state-of-the-art 53.7\% mAP, 0.4\% higher than YOLOv11-L, with superior object localization and background suppression as illustrated in Figure~\ref{fig:yolo_attention}.

In our implementation, YOLOv12 performed zero-shot detection on initial video frames within approximately 15 seconds, generating bounding boxes for all detected objects; however, filtering is often necessary to exclude non-target items (like water troughs or feeding equipment commonly present in environments such as dairy barns).

\begin{figure}[!htbp]
    \centering
    \includegraphics[width=\textwidth]{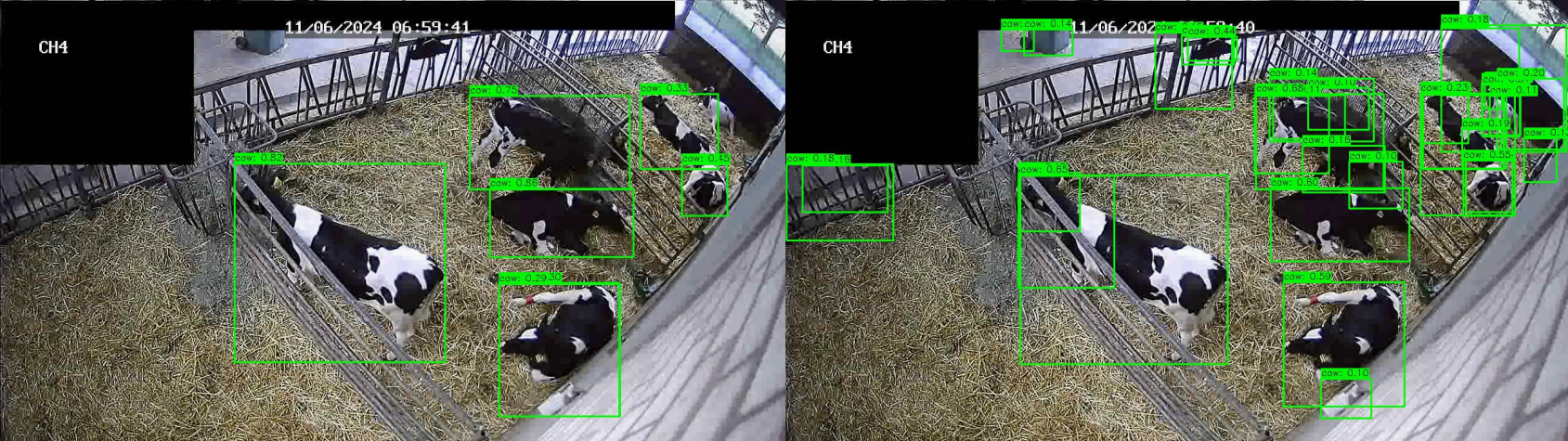}
    \caption{Bounding boxes overlaid with predictions from YOLOv12 (left) showing successful detections of cows within the frame, and OWLv2 \citep{minderer2023} detection output (right) demonstrating false positive generation at default confidence settings.}
    \label{fig:cattle_detection}
\end{figure}

We also experimented with other zero-shot object detection and localization models such as OWLv2 \citep{minderer2023}, but on the cattle footage its performance was not as good as YOLO's, as shown in Figure~\ref{fig:cattle_detection}. OWLv2 produced more false predictions and required additional adjustments of the confidence threshold to isolate useful detections.

However, the relative performance of the two models reverses when applied to the Edinburgh Pig Behavior Video from \citet{bergamini2021}.

\begin{figure}[!htbp]
    \centering
    \includegraphics[width=\textwidth]{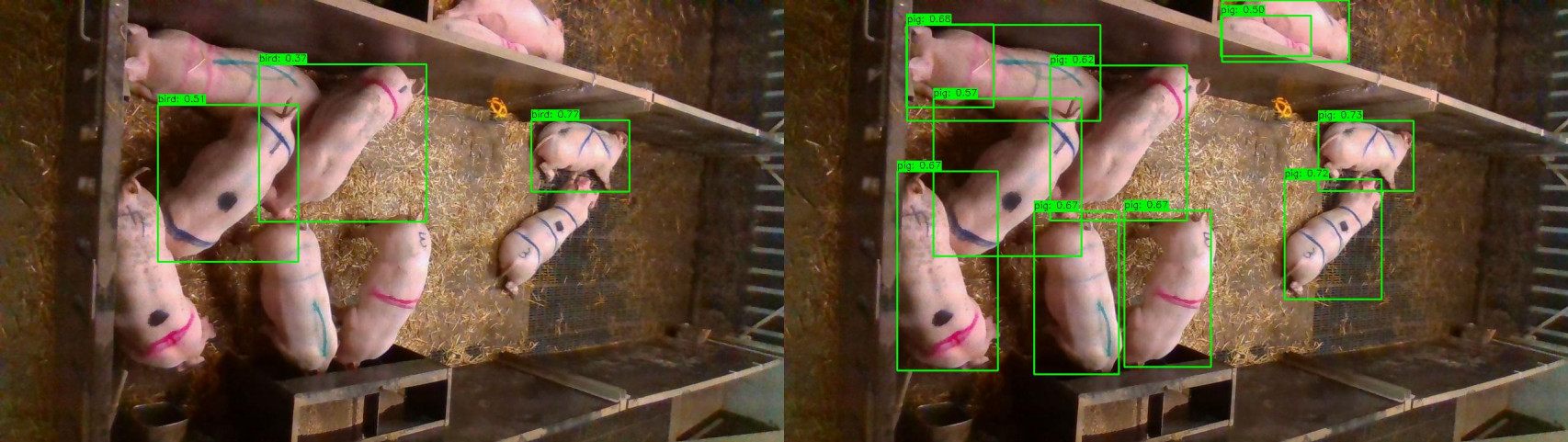}
    \caption{Detection results on overhead pig housing footage showing YOLOv12's underdetection (left) and OWLv2's complete coverage (right) at 0.5 confidence threshold. YOLOv12 detected three of the eight pigs visible in the frame, demonstrating substantial underdetection. In contrast, OWLv2 achieved complete coverage by identifying all eight target pigs and even two additional, highly occluded pigs from an adjacent pen.}
    \label{fig:pig_detection}
\end{figure}

\subsection{Segmentation and Tracking}

We also experimented with other video segmentation and tracking models such as XMem \citep{cheng2022}, but its performance was not as strong and may struggle in deployment within complex barn environments.

\begin{figure}[!htbp]
    \centering
    \includegraphics[width=\textwidth]{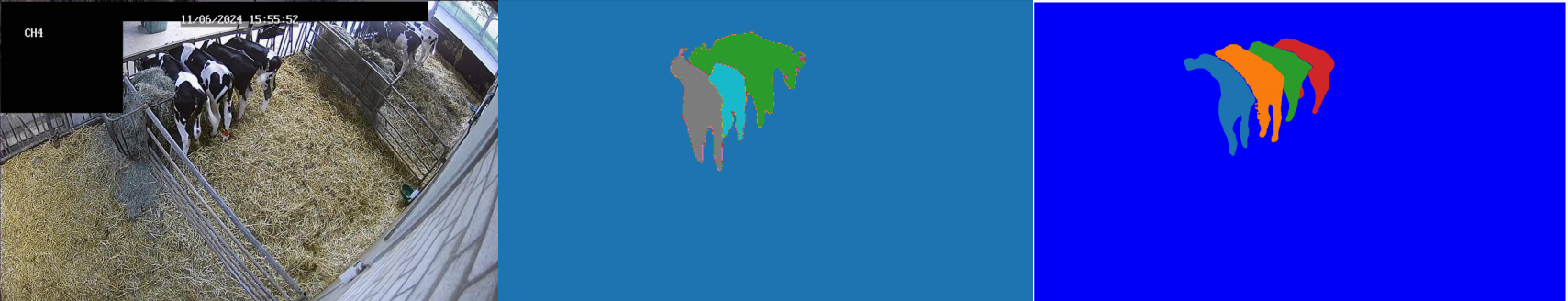}
    \caption{Comparative segmentation performance on occluded cattle showing SAMURAI's precise instance separation (middle) versus XMem's mask merging (right). The original frame is shown on the left.}
    \label{fig:segmentation}
\end{figure}

With the SAM\,2 video predictor (Figure~\ref{fig:segmentation}, middle), the cows are clearly separated, whereas the XMem \citep{cheng2022} video predictor generated a single mask covering three overlapping cows, failing to produce precise mask predictions in challenging scenes.

\subsection{Choice of Individual Modules}

For the object detection and localization component, initial experiments with YOLOv12 demonstrated promising efficiency and accuracy characteristics (see Figure~\ref{fig:pig_detection}); however, the model failed to generate reliable predictions when applied to pig detection tasks. Consequently, we adopted OWLv2 as our primary object detection and localization model. OWLv2 exhibited robust performance in our pig detection experiments, demonstrating its suitability for this specific agricultural application.

The selection of the segmentation and tracking module was based on comparative evaluation between SAMURAI and XMem models. SAMURAI demonstrated superior segmentation capabilities, producing more precise object boundaries and maintaining clearer separation between individual animals, particularly in challenging scenarios involving overlapping subjects (Figure~\ref{fig:segmentation}). The model's ability to preserve distinct contours even during close animal interactions proved critical for accurate individual tracking. These performance advantages led to the selection of SAMURAI as the segmentation and tracking component within our pipeline.

\begin{figure}[!htbp]
    \centering
    \includegraphics[width=\textwidth]{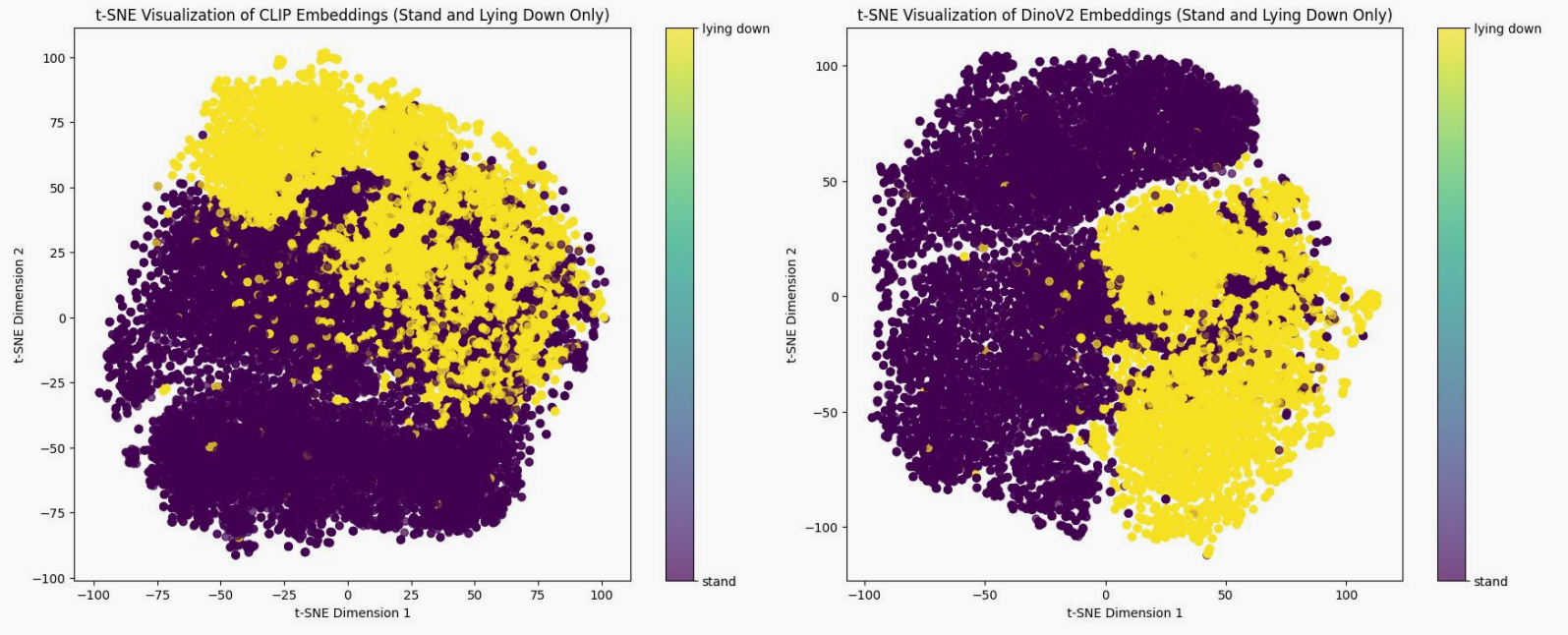}
    \caption{t-SNE visualization of CLIP embeddings (left) and DINOv2 embeddings (right) for cattle behaviors (standing vs.\ lying down). DINOv2 shows distinct cluster separation, while CLIP embeddings reveal overlapping behavioral clusters.}
    \label{fig:clip_vs_dinov2}
\end{figure}

For the feature extraction backbone, we evaluated both DINOv2 and CLIP models to determine their suitability for visual behavior analysis. As the current application does not require cross-modal text-image understanding, and given that DINOv2 demonstrated marginally superior performance in visual feature discrimination (Figure~\ref{fig:clip_vs_dinov2}), we selected DINOv2 as our feature extraction model. The t-SNE visualizations revealed more distinct behavioral clusters with DINOv2 embeddings, suggesting enhanced discriminative power for behavior classification tasks.

\section{Validation on CBVD-5}

\subsection{CBVD-5 Dataset Description}

The CBVD-5 dataset from \citet{li2024} served as a collection of dairy cow behavior keyframes with corresponding annotations, enabling us to evaluate the cross-species generalizability of our framework. The dataset originates from continuous video surveillance of dairy cows' daily activities on a commercial farm, with cameras operating 24 hours per day over a five-day period. This extensive monitoring period ensured comprehensive coverage of various behaviors across different times of day and environmental conditions.

From continuous video footage, keyframes were systematically selected for behavior annotation, resulting in 27{,}501 valid labeled data points. Each annotation contained precise spatial information about individual cow locations paired with one of five behavioral categories: standing, lying down, feeding, drinking, and rumination, capturing the primary activities essential for monitoring dairy cow welfare and productivity.

Given that the dataset provided sparse annotations across multiple cows without persistent individual identifiers, it presents different evaluation opportunities compared to the pig dataset. Consequently, we do not utilize this dataset for benchmarking our object detection, localization, tracking, and segmentation modules, which require consistent individual identification across frames. Instead, the primary evaluation focus centers on assessing the feature extraction capabilities of our framework and the classifier's ability to differentiate between behavioral patterns based on these extracted features.

For our evaluation, we focused on the standing and lying-down categories, which provide a basic test of the feature extraction capability of our pipeline. This decision was motivated by two factors: the subtle nature of rumination behavior requires exceptionally high-quality video for accurate assessment, and the feeding and drinking classes exhibited significant sample imbalance that could bias the evaluation results. After filtering to these two primary behaviors, we obtained 30{,}391 individual cropped samples suitable for analysis, providing a substantial dataset for evaluating behavioral classification performance in dairy cattle.

\subsection{Dataset Preparation}

We applied our automated cropping module to extract individual animal images from the CBVD-5 dataset annotations using the bounding boxes provided along with the pictures. The existing ground truth bounding boxes were utilized directly.

\subsection{CLIP Implementation}

CLIP \citep{radford2021} learns joint representations of images and text by training on hundreds of millions of image--text pairs scraped from the internet. The model employs two parallel encoders---a Vision Transformer for images and a Transformer for text---that are trained simultaneously to produce embeddings in a shared multimodal space.

The training objective uses a contrastive loss that maximizes the cosine similarity between matching image--text pairs while minimizing similarity between non-matching pairs. Given a batch of $N$ image--text pairs, CLIP computes pairwise cosine similarities:
\begin{equation}
S_{ij} = \frac{I_i \cdot T_j}{\|I_i\| \, \|T_j\|}
\end{equation}
where $I_i$ and $T_j$ denote the image and text embeddings, respectively.

These similarities are scaled with a temperature parameter $\tau$:
\begin{equation}
L_{ij} = \tau \cdot S_{ij}
\end{equation}

CLIP optimizes the model to maximize the similarity scores for matching pairs while minimizing scores for non-matching pairs. The image-to-text loss is:
\begin{equation}
\mathcal{L}_i = -\log \frac{\exp(L_{ii})}{\sum_{j=1}^{N} \exp(L_{ij})}
\end{equation}

The text-to-image loss is:
\begin{equation}
\mathcal{L}_t = -\log \frac{\exp(L_{ii})}{\sum_{j=1}^{N} \exp(L_{ji})}
\end{equation}

The total loss is the average of both directions:
\begin{equation}
\mathcal{L} = \tfrac{1}{2}\left(\mathcal{L}_i + \mathcal{L}_t\right)
\end{equation}

This bidirectional loss ensures that both the image and text encoders learn complementary representations. CLIP can perform zero-shot classification by comparing an input image to text descriptions of different classes. The model computes the similarity between the image embedding and text embeddings for phrases like ``a photo of a [class]'' and predicts the class with the highest similarity score.

Computing the image embedding:
\begin{equation}
I_{\text{test}} = I(\text{test image})
\end{equation}

Computing text embeddings for each class:
\begin{equation}
T_c = T(\text{``a photo of a \{class\}''})
\end{equation}

Predicting the class with maximum similarity:
\begin{equation}
\hat{y} = \arg\max_{c} \frac{I_{\text{test}} \cdot T_c}{\|I_{\text{test}}\| \, \|T_c\|}
\end{equation}

This ability to generalize to new tasks without additional training makes CLIP particularly versatile for various computer vision applications. To get more versatile features, CLIP-ViT-Large-Patch14 was selected for its strong zero-shot accuracy.

The implementation followed a similar pipeline to DINOv2 but includes additional capabilities for text-based queries, enabling zero-shot behavior classification without fine-tuning. However, in our experiments, we used these models primarily for feature extraction rather than zero-shot classification, as we trained supervised classifiers on the extracted embeddings.

\subsection{Feature Extraction Comparison on Cows}

We compared DINOv2 and CLIP embeddings using t-SNE visualization to evaluate their discriminative power for behavior classification.

The visualizations reveal that DINOv2 produces more distinct clusters for different behaviors, while CLIP embeddings show moderate separation with overlapping regions.

\subsection{Classification of Cow Behaviors}

We evaluated MLP classifiers on the extracted features from both DINOv2 and CLIP models to assess their discriminative power for behavior classification, with training typically converging within 3--5 epochs.

\begin{table}[!htbp]
    \centering
    \caption{Binary classification performance of MLP models using DINOv2 and CLIP features for standing versus lying behavior detection of dairy cows.}
    \label{tab:cow_overall}
    \begin{tabular}{lccccc}
        \toprule
        \textbf{Model} & \textbf{Feature Type} & \textbf{Accuracy} & \textbf{Precision} & \textbf{Recall} & \textbf{F1-Score} \\
        \midrule
        MLP & DINOv2 & 98.3\% & 0.982 & 0.982 & 0.982 \\
        MLP & CLIP   & 98.2\% & 0.981 & 0.983 & 0.982 \\
        \bottomrule
    \end{tabular}
\end{table}

Both feature extraction methods achieved high performance, with nearly identical results. The MLP classifier using DINOv2 features achieved 98.3\% accuracy with a test loss of 0.0645, while the CLIP-based classifier achieved 98.2\% accuracy with a slightly lower test loss of 0.0565.

Detailed performance metrics by class reveal consistent performance across both behaviors (Table~\ref{tab:cow_per_class}).

\begin{table}[!htbp]
    \centering
    \caption{Class-specific performance metrics for standing and lying behavior classification using DINOv2 and CLIP features.}
    \label{tab:cow_per_class}
    \begin{tabular}{llcccc}
        \toprule
        \textbf{Feature Extractor} & \textbf{Behavior} & \textbf{Precision} & \textbf{Recall} & \textbf{F1-Score} & \textbf{Support} \\
        \midrule
        \multirow{2}{*}{DINOv2}
            & Standing & 0.987 & 0.984 & 0.985 & 3022 \\
            & Lying    & 0.977 & 0.980 & 0.978 & 2042 \\
        \midrule
        \multirow{2}{*}{CLIP}
            & Standing & 0.990 & 0.980 & 0.985 & 3022 \\
            & Lying    & 0.971 & 0.985 & 0.978 & 2042 \\
        \bottomrule
    \end{tabular}
\end{table}

The high performance of both feature extraction methods indicates that our framework successfully captures discriminative features for behavior classification. The marginal difference between DINOv2 and CLIP suggests that both approaches effectively encode behavioral information, with DINOv2 showing a slight edge in overall accuracy. The balanced performance across classes demonstrates the robustness of our approach for identifying both standing and lying behaviors with minimal bias toward either category.

\section{Glossary}

\begin{description}[leftmargin=!,labelwidth=2.5cm,style=nextline,topsep=2pt,itemsep=0pt]
    \item[A2] Area Attention
    \item[AI] Artificial Intelligence
    \item[AO] Average Overlap
    \item[API] Application Programming Interface
    \item[AUC] Area Under the Curve
    \item[CBVD-5] Cow Behavior Video Dataset (5 categories)
    \item[CLS] Class token (in transformers)
    \item[CLIP] Contrastive Language-Image Pre-training
    \item[CNN] Convolutional Neural Network
    \item[CPU] Central Processing Unit
    \item[DINOv2] Self-Distillation with NO labels, version 2
    \item[GB] Gigabyte
    \item[GOT-10k] Generic Object Tracking benchmark (10,000 videos)
    \item[GPU] Graphics Processing Unit
    \item[I/O] Input/Output
    \item[IoU] Intersection over Union
    \item[LaSOT] Large-scale Single Object Tracking
    \item[LaSOText] LaSOT extension dataset
    \item[LSTM] Long Short-Term Memory
    \item[LVD-142M] Large Vision Dataset with 142 Million images
    \item[mAP] mean Average Precision
    \item[MLP] Multi-Layer Perceptron
    \item[NVMe] Non-Volatile Memory Express
    \item[OWLv2] Open-World Localization Vision model, version 2
    \item[$P_{\text{norm}}$] Normalized Precision
    \item[RAM] Random Access Memory
    \item[ReLU] Rectified Linear Unit
    \item[RGB] Red, Green, Blue (color model)
    \item[SAM] Segment Anything Model
    \item[SAM\,2] Segment Anything Model, version 2
    \item[SAMURAI] Segment Anything Model Upgraded for Real-time Adaptation and Inference
    \item[SSD] Solid State Drive
    \item[t-SNE] t-distributed Stochastic Neighbor Embedding
    \item[V100] NVIDIA Volta 100 GPU
    \item[ViT] Vision Transformer
    \item[XMem] Extended Memory (video segmentation model)
    \item[YOLO] You Only Look Once
\end{description}

\bibliographystyle{plainnat}
\bibliography{references}

@article{alvarenga2021,
  author    = {Alvarenga, A. B. and Oliveira, H. R. and Chen, S. Y. and Miller, S. P. and Marchant-Forde, J. N. and Grigoletto, L. and Brito, L. F.},
  title     = {A systematic review of genomic regions and candidate genes underlying behavioral traits in farmed mammals and their link with human disorders},
  journal   = {Animals},
  volume    = {11},
  number    = {3},
  pages     = {715},
  year      = {2021}
}

@article{antanaitis2023,
  author    = {Antanaitis, R. and D{\v{z}}ermeikait{\.{e}}, K. and Bespalovait{\.{e}}, A. and Ribelyt{\.{e}}, I. and Rutkauskas, A. and Japertas, S. and Baumgartner, W.},
  title     = {Assessment of ruminating, eating, and locomotion behavior during heat stress in dairy cattle by using advanced technological monitoring},
  journal   = {Animals},
  volume    = {13},
  number    = {18},
  pages     = {2825},
  year      = {2023}
}

@article{arablouei2023,
  author    = {Arablouei, R. and Wang, Z. and Bishop-Hurley, G. J. and Liu, J.},
  title     = {Multimodal sensor data fusion for in-situ classification of animal behavior using accelerometry and {GNSS} data},
  journal   = {Smart Agricultural Technology},
  volume    = {4},
  pages     = {100163},
  year      = {2023}
}

@book{bass2021,
  author    = {Bass, L. and Clements, P. C. and Kazman, R.},
  title     = {Software architecture in practice},
  edition   = {4},
  publisher = {Addison-Wesley Professional},
  year      = {2021}
}

@article{berckmans2014,
  author    = {Berckmans, D.},
  title     = {Precision livestock farming technologies for welfare management in intensive livestock systems},
  journal   = {Revue Scientifique et Technique},
  volume    = {33},
  number    = {1},
  pages     = {189--196},
  year      = {2014}
}

@inproceedings{bergamini2021,
  author    = {Bergamini, L. and Pini, S. and Simoni, A. and Vezzani, R. and Calderara, S. and Eath, R. B. and Fisher, R. B.},
  title     = {Extracting accurate long-term behavior changes from a large pig dataset},
  booktitle = {16th International Joint Conference on Computer Vision, Imaging and Computer Graphics Theory and Applications (VISIGRAPP)},
  pages     = {524--533},
  publisher = {SciTePress},
  year      = {2021}
}

@article{bokade2021,
  author    = {Bokade, R. and Navato, A. and Ouyang, R. and Jin, X. and Chou, C. A. and Ostadabbas, S. and Mueller, A. V.},
  title     = {A cross-disciplinary comparison of multimodal data fusion approaches and applications: Accelerating learning through trans-disciplinary information sharing},
  journal   = {Expert Systems with Applications},
  volume    = {165},
  pages     = {113885},
  year      = {2021}
}

@article{bonneau2020,
  author    = {Bonneau, M. and Vayssade, J. A. and Troupe, W. and Arquet, R.},
  title     = {Outdoor animal tracking combining neural network and time-lapse cameras},
  journal   = {Computers and Electronics in Agriculture},
  volume    = {168},
  pages     = {105150},
  year      = {2020}
}

@article{bugueiro2021,
  author    = {Bugueiro, A. and Fouz, R. and Di{\'e}guez, F. J.},
  title     = {Associations between on-farm welfare, milk production, and reproductive performance in dairy herds in Northwestern Spain},
  journal   = {Journal of Applied Animal Welfare Science},
  volume    = {24},
  number    = {1},
  pages     = {29--38},
  year      = {2021}
}

@inproceedings{cheng2022,
  author    = {Cheng, H. K. and Schwing, A. G.},
  title     = {{XMem}: Long-term video object segmentation with an {Atkinson--Shiffrin} memory model},
  booktitle = {European Conference on Computer Vision (ECCV)},
  pages     = {640--658},
  publisher = {Springer Nature Switzerland},
  year      = {2022}
}

@article{clark2019,
  author    = {Clark, B. and Panzone, L. A. and Stewart, G. B. and Kyriazakis, I. and Niemi, J. K. and Latvala, T. and Frewer, L. J.},
  title     = {Consumer attitudes towards production diseases in intensive production systems},
  journal   = {PLoS One},
  volume    = {14},
  number    = {1},
  pages     = {e0210432},
  year      = {2019}
}

@article{cornish2016,
  author    = {Cornish, A. and Raubenheimer, D. and McGreevy, P.},
  title     = {What we know about the public's level of concern for farm animal welfare in food production in developed countries},
  journal   = {Animals},
  volume    = {6},
  number    = {11},
  pages     = {74},
  year      = {2016}
}

@article{domun2019,
  author    = {Domun, Y. and Pedersen, L. J. and White, D. and Adeyemi, O. and Norton, T.},
  title     = {Learning patterns from time-series data to discriminate predictions of tail-biting, fouling and diarrhoea in pigs},
  journal   = {Computers and Electronics in Agriculture},
  volume    = {163},
  pages     = {104878},
  year      = {2019}
}

@article{dzermeikaite2023,
  author    = {D{\v{z}}ermeikait{\.{e}}, K. and Ba{\v{c}}{\.{e}}ninait{\.{e}}, D. and Antanaitis, R.},
  title     = {Innovations in cattle farming: application of innovative technologies and sensors in the diagnosis of diseases},
  journal   = {Animals},
  volume    = {13},
  number    = {5},
  pages     = {780},
  year      = {2023}
}

@article{eckelkamp2020,
  author    = {Eckelkamp, E. A. and Bewley, J. M.},
  title     = {On-farm use of disease alerts generated by precision dairy technology},
  journal   = {Journal of Dairy Science},
  volume    = {103},
  number    = {2},
  pages     = {1566--1582},
  year      = {2020}
}

@article{fan2021,
  author    = {Fan, H. and Bai, H. and Lin, L. and Yang, F. and Chu, P. and Deng, G. and Ling, H.},
  title     = {{LaSOT}: A high-quality large-scale single object tracking benchmark},
  journal   = {International Journal of Computer Vision},
  volume    = {129},
  pages     = {439--461},
  year      = {2021}
}

@inproceedings{fan2019,
  author    = {Fan, H. and Lin, L. and Yang, F. and Chu, P. and Deng, G. and Yu, S. and Ling, H.},
  title     = {{LaSOT}: A high-quality benchmark for large-scale single object tracking},
  booktitle = {Proceedings of the IEEE/CVF Conference on Computer Vision and Pattern Recognition (CVPR)},
  pages     = {5374--5383},
  year      = {2019}
}

@article{fernandes2020,
  author    = {Fernandes, A. F. A. and D{\'o}rea, J. R. R. and Rosa, G. J. D. M.},
  title     = {Image analysis and computer vision applications in animal sciences: an overview},
  journal   = {Frontiers in Veterinary Science},
  volume    = {7},
  pages     = {551269},
  year      = {2020}
}

@article{firk2002,
  author    = {Firk, R. and Stamer, E. and Junge, W. and Krieter, J.},
  title     = {Automation of oestrus detection in dairy cows: a review},
  journal   = {Livestock Production Science},
  volume    = {75},
  number    = {3},
  pages     = {219--232},
  year      = {2002}
}

@article{fuentes2023,
  author    = {Fuentes, A. and Han, S. and Nasir, M. F. and Park, J. and Yoon, S. and Park, D. S.},
  title     = {Multiview monitoring of individual cattle behavior based on action recognition in closed barns using deep learning},
  journal   = {Animals},
  volume    = {13},
  number    = {12},
  pages     = {2020},
  year      = {2023}
}

@article{guo2023,
  author    = {Guo, Q. and Sun, Y. and Orsini, C. and Bolhuis, J. E. and de Vlieg, J. and Bijma, P. and de With, P. H.},
  title     = {Enhanced camera-based individual pig detection and tracking for smart pig farms},
  journal   = {Computers and Electronics in Agriculture},
  volume    = {211},
  pages     = {108009},
  year      = {2023}
}

@article{halachmi2019,
  author    = {Halachmi, I. and Guarino, M. and Bewley, J. and Pastell, M.},
  title     = {Smart animal agriculture: application of real-time sensors to improve animal well-being and production},
  journal   = {Annual Review of Animal Biosciences},
  volume    = {7},
  number    = {1},
  pages     = {403--425},
  year      = {2019}
}

@article{hampton2019,
  author    = {Hampton, J. O. and MacKenzie, D. I. and Forsyth, D. M.},
  title     = {How many to sample? Statistical guidelines for monitoring animal welfare outcomes},
  journal   = {PLoS One},
  volume    = {14},
  number    = {1},
  pages     = {e0211417},
  year      = {2019}
}

@article{huang2019,
  author    = {Huang, L. and Zhao, X. and Huang, K.},
  title     = {{GOT-10k}: A large high-diversity benchmark for generic object tracking in the wild},
  journal   = {IEEE Transactions on Pattern Analysis and Machine Intelligence},
  volume    = {43},
  number    = {5},
  pages     = {1562--1577},
  year      = {2019}
}

@inproceedings{kingma2014,
  author    = {Kingma, D. P. and Ba, J.},
  title     = {{Adam}: A method for stochastic optimization},
  booktitle = {International Conference on Learning Representations (ICLR)},
  year      = {2015}
}

@article{kirillov2023,
  author    = {Kirillov, A. and Mintun, E. and Ravi, N. and Mao, H. and Rolland, C. and Gustafson, L. and Xiao, T. and Whitehead, S. and Berg, A. C. and Lo, W.-Y. and Doll{\'a}r, P. and Girshick, R.},
  title     = {Segment Anything},
  journal   = {arXiv preprint arXiv:2304.02643},
  year      = {2023}
}

@article{li2021,
  author    = {Li, G. and Huang, Y. and Chen, Z. and Chesser, G. D. and Purswell, J. L. and Linhoss, J. and Zhao, Y.},
  title     = {Practices and applications of convolutional neural network-based computer vision systems in animal farming: A review},
  journal   = {Sensors},
  volume    = {21},
  number    = {4},
  pages     = {1492},
  year      = {2021}
}

@article{li2024,
  author    = {Li, K. and Fan, D. and Wu, H. and Zhao, A.},
  title     = {A new dataset for video-based cow behavior recognition},
  journal   = {Scientific Reports},
  volume    = {14},
  number    = {1},
  pages     = {18702},
  year      = {2024}
}

@article{liu2024,
  author    = {Liu, Y. and Li, W. and Liu, X. and Li, Z. and Yue, J.},
  title     = {Deep learning in multiple animal tracking: A survey},
  journal   = {Computers and Electronics in Agriculture},
  volume    = {224},
  pages     = {109161},
  year      = {2024}
}

@article{mathis2018,
  author    = {Mathis, A. and Mamidanna, P. and Cury, K. M. and Abe, T. and Murthy, V. N. and Mathis, M. W. and Bethge, M.},
  title     = {{DeepLabCut}: markerless pose estimation of user-defined body parts with deep learning},
  journal   = {Nature Neuroscience},
  volume    = {21},
  number    = {9},
  pages     = {1281--1289},
  year      = {2018}
}

@article{matthews2016,
  author    = {Matthews, S. G. and Miller, A. L. and Clapp, J. and Pl{\"o}tz, T. and Kyriazakis, I.},
  title     = {Early detection of health and welfare compromises through automated detection of behavioural changes in pigs},
  journal   = {The Veterinary Journal},
  volume    = {217},
  pages     = {43--51},
  year      = {2016}
}

@article{matthews2017,
  author    = {Matthews, S. G. and Miller, A. L. and Pl{\"o}tz, T. and Kyriazakis, I.},
  title     = {Automated tracking to measure behavioural changes in pigs for health and welfare monitoring},
  journal   = {Scientific Reports},
  volume    = {7},
  number    = {1},
  pages     = {17582},
  year      = {2017}
}

@article{mcdonagh2021,
  author    = {McDonagh, J. and Tzimiropoulos, G. and Slinger, K. R. and Huggett, Z. J. and Down, P. M. and Bell, M. J.},
  title     = {Detecting dairy cow behavior using vision technology},
  journal   = {Agriculture},
  volume    = {11},
  number    = {7},
  pages     = {675},
  year      = {2021}
}

@article{menezes2024,
  author    = {Menezes, G. L. and Mazon, G. and Ferreira, R. E. and Cabrera, V. E. and Dorea, J. R.},
  title     = {Artificial intelligence for livestock: a narrative review of the applications of computer vision systems and large language models for animal farming},
  journal   = {Animal Frontiers},
  volume    = {14},
  number    = {6},
  pages     = {42--53},
  year      = {2024}
}

@article{minderer2023,
  author    = {Minderer, M. and Gritsenko, A. and Houlsby, N.},
  title     = {Scaling open-vocabulary object detection},
  journal   = {Advances in Neural Information Processing Systems},
  volume    = {36},
  pages     = {72983--73007},
  year      = {2023}
}

@inproceedings{minderer2022,
  author    = {Minderer, M. and Gritsenko, A. and Stone, A. and Neumann, M. and Weissenborn, D. and Dosovitskiy, A. and Houlsby, N.},
  title     = {Simple open-vocabulary object detection},
  booktitle = {European Conference on Computer Vision (ECCV)},
  pages     = {728--755},
  publisher = {Springer Nature Switzerland},
  year      = {2022}
}

@article{neethirajan2021,
  author    = {Neethirajan, S.},
  title     = {The use of artificial intelligence in assessing affective states in livestock},
  journal   = {Frontiers in Veterinary Science},
  volume    = {8},
  pages     = {715261},
  year      = {2021}
}

@article{oliveira2021,
  author    = {Oliveira, D. A. B. and Pereira, L. G. R. and Bresolin, T. and Ferreira, R. E. P. and Dorea, J. R. R.},
  title     = {A review of deep learning algorithms for computer vision systems in livestock},
  journal   = {Livestock Science},
  volume    = {253},
  pages     = {104700},
  year      = {2021}
}

@article{oquab2023,
  author    = {Oquab, M. and Darcet, T. and Moutakanni, T. and Vo, H. and Szafraniec, M. and Khalidov, V. and Bojanowski, P.},
  title     = {{DINOv2}: Learning robust visual features without supervision},
  journal   = {arXiv preprint arXiv:2304.07193},
  year      = {2023}
}

@inproceedings{orhei2020,
  author    = {Orhei, C. and Mocofan, M. and Vert, S. and Vasiu, R.},
  title     = {End-to-end computer vision framework},
  booktitle = {2020 International Symposium on Electronics and Telecommunications (ISETC)},
  pages     = {1--4},
  publisher = {IEEE},
  year      = {2020}
}

@article{psota2020,
  author    = {Psota, E. T. and Schmidt, T. and Mote, B. and P{\'e}rez, L. C.},
  title     = {Long-term tracking of group-housed livestock using keypoint detection and {MAP} estimation for individual animal identification},
  journal   = {Sensors},
  volume    = {20},
  number    = {13},
  pages     = {3670},
  year      = {2020}
}

@inproceedings{radford2021,
  author    = {Radford, A. and Kim, J. W. and Hallacy, C. and Ramesh, A. and Goh, G. and Agarwal, S. and Sastry, G. and Askell, A. and Mishkin, P. and Clark, J. and Krueger, G. and Sutskever, I.},
  title     = {Learning transferable visual models from natural language supervision},
  booktitle = {International Conference on Machine Learning (ICML)},
  pages     = {8748--8763},
  publisher = {PMLR},
  year      = {2021}
}

@article{ravi2024,
  author    = {Ravi, N. and Gabeur, V. and Hu, Y. T. and Hu, R. and Ryali, C. and Ma, T. and Feichtenhofer, C.},
  title     = {{SAM 2}: Segment anything in images and videos},
  journal   = {arXiv preprint arXiv:2408.00714},
  year      = {2024}
}

@article{rial2024,
  author    = {Rial, C. and Stangaferro, M. L. and Thomas, M. J. and Giordano, J. O.},
  title     = {Effect of automated health monitoring based on rumination, activity, and milk yield alerts versus visual observation on herd health monitoring and performance outcomes},
  journal   = {Journal of Dairy Science},
  volume    = {107},
  number    = {12},
  pages     = {11576--11596},
  year      = {2024}
}

@article{rohan2024,
  author    = {Rohan, A. and Rafaq, M. S. and Hasan, M. J. and Asghar, F. and Bashir, A. K. and Dottorini, T.},
  title     = {Application of deep learning for livestock behaviour recognition: A systematic literature review},
  journal   = {Computers and Electronics in Agriculture},
  volume    = {224},
  pages     = {109115},
  year      = {2024}
}

@article{rumelhart1986,
  author    = {Rumelhart, D. E. and Hinton, G. E. and Williams, R. J.},
  title     = {Learning representations by back-propagating errors},
  journal   = {Nature},
  volume    = {323},
  number    = {6088},
  pages     = {533--536},
  year      = {1986}
}

@article{stygar2021,
  author    = {Stygar, A. H. and G{\'o}mez, Y. and Berteselli, G. V. and Dalla Costa, E. and Canali, E. and Niemi, J. K. and Pastell, M.},
  title     = {A systematic review on commercially available and validated sensor technologies for welfare assessment of dairy cattle},
  journal   = {Frontiers in Veterinary Science},
  volume    = {8},
  pages     = {634338},
  year      = {2021}
}

@article{tian2020,
  author    = {Tian, Hongkun and Wang, Tianhai and Liu, Yadong and Qiao, Xi and Li, Yanzhou},
  title     = {Computer vision technology in agricultural automation---A review},
  journal   = {Information Processing in Agriculture},
  volume    = {7},
  number    = {1},
  pages     = {1--19},
  year      = {2020}
}

@article{tian2025,
  author    = {Tian, Y. and Ye, Q. and Doermann, D.},
  title     = {{YOLOv12}: Attention-centric real-time object detectors},
  journal   = {arXiv preprint arXiv:2502.12524},
  year      = {2025}
}

@article{vidal2021,
  author    = {Vidal, M. and Wolf, N. and Rosenberg, B. and Harris, B. P. and Mathis, A.},
  title     = {Perspectives on individual animal identification from biology and computer vision},
  journal   = {Integrative and Comparative Biology},
  volume    = {61},
  number    = {3},
  pages     = {900--916},
  year      = {2021}
}

@article{wang2024,
  author    = {Wang, X. and Wang, Y. and Yang, J. and Jia, X. and Li, L. and Ding, W. and Wang, F. Y.},
  title     = {The survey on multi-source data fusion in cyber-physical-social systems: Foundational infrastructure for industrial metaverses and industries 5.0},
  journal   = {Information Fusion},
  pages     = {102321},
  year      = {2024}
}

@article{wurtz2019,
  author    = {Wurtz, K. and Camerlink, I. and D'Eath, R. B. and Fern{\'a}ndez, A. P. and Norton, T. and Steibel, J. and Siegford, J.},
  title     = {Recording behaviour of indoor-housed farm animals automatically using machine vision technology: A systematic review},
  journal   = {PLoS One},
  volume    = {14},
  number    = {12},
  pages     = {e0226669},
  year      = {2019}
}

@article{yang2024,
  author    = {Yang, C. Y. and Huang, H. W. and Chai, W. and Jiang, Z. and Hwang, J. N.},
  title     = {{SAMURAI}: Adapting segment anything model for zero-shot visual tracking with motion-aware memory},
  journal   = {arXiv preprint arXiv:2411.11922},
  year      = {2024}
}

\end{document}